\documentclass{article}

\usepackage{xcolor}
\usepackage[preprint]{corl_2026}
\hypersetup{
  pdftitle={The Relative Value of Demonstration Data: Execution Time Limits and Operator Mixtures in Behavior Cloning},
  pdfauthor={Jiaxuan Guo, Xingyu Ren, Youran Sun, Jingxin Yang, Kejia Zhang, Haizhao Yang},
  pdfsubject={},
  pdfkeywords={imitation learning, data scaling, demonstration quality, teleoperation}}

\usepackage[T1]{fontenc}
\usepackage{amsmath}
\usepackage{amssymb}
\usepackage{graphicx}
\usepackage{booktabs}
\usepackage{array}
\usepackage{placeins}
\usepackage{microtype}
\usepackage{enumitem}
\usepackage[capitalize,nameinlink]{cleveref}

\newcounter{infobox}
\crefname{infobox}{Box}{Boxes}
\Crefname{infobox}{Box}{Boxes}
\newlist{findings}{enumerate}{1}
\setlist[findings]{label=(\roman*), leftmargin=*, topsep=2pt, itemsep=1pt}

\newcommand{\kadd}{k_{\mathrm{add}}}
\newcommand{\Square}{\textit{Square}}
\newcommand{\Can}{\textit{Can}}
\newcommand{\Better}{\textit{Better}}
\newcommand{\Worse}{\textit{Worse}}
\newcommand{\Okay}{\textit{Okay}}

\title{The Relative Value of Demonstration Data:\\ Execution Time Limits and Operator Mixtures\\ in Behavior Cloning}

\author{%
  Jiaxuan Guo$^{1,*}$\enspace
  Xingyu Ren$^{2}$\enspace
  Youran Sun$^{3}$\enspace
  Jingxin Yang$^{1}$\enspace
  Kejia Zhang$^{3}$\enspace
  Haizhao Yang$^{3}$\\
  {\normalfont $^{1}$Stanford University\quad
  $^{2}$The Chinese University of Hong Kong}\\
  {\normalfont $^{3}$University of Maryland, College Park}\\
  {\normalfont $^{*}$Corresponding author: \texttt{guojx@stanford.edu}}%
}

\begin{document}
\maketitle
\enlargethispage{-8.05034pt}
\suppressfloats[t]

\begin{abstract}
Robot demonstration datasets combine demonstrations from teleoperators of uneven experience, so collection teams need to know how many demonstrations from one group match those from another.
We measure this with a data multiple: the ratio of the numbers of demonstrations two groups need for their policies to reach the same success rate.
On robomimic's simulated \Square{} and \Can{} tasks, we train behavior-cloning policies on demonstrations from the two teleoperators robomimic labels \Better{} and the two it labels \Worse{}, varying dataset size and the share of \Worse{} demonstrations.
On \Square{}, within robomimic's 25-second time limit, an all-\Worse{} training set needs 1.55 times as many demonstrations as an all-\Better{} set.
Most of this difference is late completion: policies trained on \Worse{} demonstrations finish later, as the \Worse{} teleoperators do, and with a 100-second limit the multiple falls to 1.11.
On \Can{}, most of the difference remains at 100 seconds.
If each \Worse{} demonstration were worth 1/1.55 of a \Better{} one, as in an all-\Worse{} set, a half-\Better{}, half-\Worse{} set on \Square{} would need 1.22 times as many demonstrations as an all-\Better{} set.
In fact, it needs 13\% more than that, and this excess persists at every time limit from 20 to 200 seconds.
The excess remains when training draws demonstrations uniformly instead of in proportion to their length.
On \Square{}, within the 25-second limit, switching to these \Worse{} teleoperators is cheaper at equal success only if they earn less than a third of the \Better{} teleoperators' hourly pay, because \Worse{} demonstrations also take longer to record.
\end{abstract}

\section{Introduction}
\label{sec:intro}

\begin{figure}[t]
  \centering
  \includegraphics[width=\linewidth]{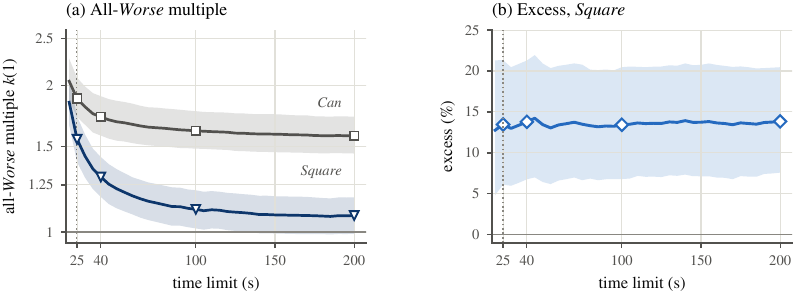}
  \caption{\textbf{A longer time limit shrinks the all-\Worse{} multiple on \Square{}, and the excess persists.}
  (a)~All-\Worse{} multiple $k(1)$ on \Square{} and \Can{} (log scale): how many times as many demonstrations an all-\Worse{} set needs as an all-\Better{} set for the same success rate.
  (b)~Excess of the half-\Better{}, half-\Worse{} set on \Square{}, in percent beyond its additive prediction.
  Each policy's final checkpoint is scored every 5\,s from 20\,s to 200\,s; open markers: 25, 40, 100, and 200\,s; dotted line: robomimic's limit; bands: seed intervals (\cref{sec:setup}).}
  \label{fig:horizon}
\end{figure}

Robot demonstration datasets are recorded by many teleoperators of uneven experience, so a team that plans a collection has to know how much each teleoperator's demonstrations are worth.
DROID, for example, pools demonstrations from many teleoperators~\citep{khazatsky2024droid}.
Robomimic released demonstrations from six teleoperators that it labels \Better{}, \Okay{}, and \Worse{}, two of each, and policies trained on the \Better{} teleoperators' demonstrations succeed more often~\citep{mandlekar2021matters}.
Planning a collection requires an exchange rate between such groups: how many demonstrations from one group do the work of one demonstration from another.

Robomimic compares its groups by training policies on equal numbers of demonstrations from each group and scoring them within one time limit~\citep{mandlekar2021matters}.
Two conditions of such a comparison can change the exchange rate, so both have to be measured.
The first is the time limit, because a single limit counts a success that arrives after it as a failure, and robomimic's \Worse{} demonstrations are long.
On its \Square{} task, a \Worse{} demonstration lasts 1.93 times as long as a \Better{} one on average.
Of the 100 \Worse{} demonstrations, 13 run past robomimic's time limit of 25\,s (500 steps at 20\,Hz).
The second is pooling: a set that combines the two groups can be worth more or less than their separate values predict.

We measure the exchange rate as a data multiple, following the effective-data idea of \citet{hernandez2021scaling}.
The data multiple is the ratio of the numbers of demonstrations two training sets need for their policies to reach the same success rate.
On robomimic's \Square{} and \Can{} tasks, we train behavior-cloning policies on demonstrations from the two \Better{} and the two \Worse{} teleoperators, varying the dataset size and the share of \Worse{} demonstrations.
Every evaluation episode records the time of its first success.
We can therefore score the same policies at time limits of 25, 40, 100, and 200\,s without retraining them.
To test pooling, we compare mixed sets, which contain demonstrations from both groups, with an additive prediction at a fixed size.
The additive prediction credits each demonstration with the value it has in a training set from its own group.

Robomimic released the teleoperator data we use and compared its groups by success rate within its 25\,s limit~\citep{mandlekar2021matters}.
We express the gap in demonstrations and follow it across time limits on the same trained policies.
\Citet{hernandez2021scaling} introduced effective data to measure what pre-training is worth to a language model.
We apply this accounting to groups of teleoperators and to closed-loop success, and we test whether the separately measured values add up when the groups are pooled.
Curation and weighting methods decide which demonstrations a policy trains on~\citep{hejna2025robot,hejna2025remix,gandhi2022eliciting}.
We keep the data fixed and measure what each group's demonstrations are worth, in demonstrations and in recording pay.
We also test whether pooling still costs demonstrations when training draws every demonstration equally often.

Our contributions are the following.
\begin{findings}
\item On \Square{}, most of the gap between the two groups is late completion: successes that arrive after robomimic's 25\,s limit but within a longer one (\cref{sec:speed}, \cref{fig:horizon}a). Policies trained on \Worse{} demonstrations finish later, as the \Worse{} teleoperators do. On \Can{}, a longer limit removes much less of the gap.
\item Pooling all four teleoperators costs extra demonstrations, and this cost persists at longer time limits (\cref{sec:mixing}). A half-\Better{}, half-\Worse{} set on \Square{} holds equal numbers of demonstrations from the two groups. At every time limit from 20\,s to 200\,s, it needs more demonstrations than its additive prediction (\cref{fig:horizon}b); we call the extra percentage the excess. The excess remains when training draws demonstrations uniformly instead of in proportion to their length.
\item We convert the multiples into break-even pay ratios for two collection decisions: replacing \Better{} demonstrations with \Worse{} ones, and adding \Worse{} demonstrations to a \Better{} set (\cref{sec:implications}).
\end{findings}
\section{Measuring What a Demonstration Is Worth}
\label{sec:setup}

\paragraph{Data.}
We use robomimic's multi-human datasets for two simulated tasks~\citep{mandlekar2021matters}.
In \Square{}, the robot places a square nut on a rod; in \Can{}, it moves a can from a large bin into a smaller one.
Each dataset was recorded through RoboTurk by six teleoperators, 50 demonstrations each~\citep{mandlekar2018roboturk,mandlekar2021matters}.
Robomimic labels two of them \Better{} (experienced), two \Okay{} (adequate), and two \Worse{} (inexperienced).
We use the \Better{} teleoperators B1 and B2 and the \Worse{} teleoperators W1 and W2, so each group contributes 100 demonstrations per task.
Policies observe the end-effector and gripper states and the object poses.

\paragraph{Design.}
A training set has a size $N$, its number of demonstrations, and a \Worse{} share $p$, the fraction of its demonstrations recorded by the \Worse{} teleoperators.
The all-\Better{} and all-\Worse{} sets have $p=0$ and $p=1$, and the half-\Better{}, half-\Worse{} set has $p=0.5$.
A training seed fixes both a random order of the demonstrations and the training randomness, and a seed's larger training sets contain its smaller ones.
Within a group, the two teleoperators contribute equal numbers of demonstrations, up to rounding.

\paragraph{Main experiment.}
On \Square{}, we cross eight sizes from $N=12$ to 100 with $p\in\{0, 0.25, 0.5, 0.75, 1\}$, using 24 training seeds for $p\in\{0, 0.5, 1\}$ and 12 of them for the other two shares.
On \Can{}, we cross six sizes from $N=10$ to 100 with $p\in\{0, 0.5, 1\}$ and 12 seeds.
We call these two designs the main experiment (\cref{tab:protocols}).

\paragraph{Training and evaluation.}
We train robomimic's low-dimensional recurrent behavior-cloning policy with a Gaussian-mixture action head (BC-RNN-GMM) with robomimic's settings~\citep{mandlekar2021matters}.
Robomimic's default sampler draws short training sequences uniformly over all start positions, so each demonstration is drawn in proportion to its length.
In the final-checkpoint evaluation, we run the final checkpoint of each policy from 100 initial states that all policies share.
Each episode stops at its first success or after 200\,s, and we record the time of that success.
Scoring an episode at a longer time limit thus adds the successes that arrive later in the same episode.
Under robomimic's protocol, our success rates differ from its published values by at most 0.06 and 0.07 on \Square{} and \Can{} (\cref{app:known}).

\begin{figure}[t]
\centering
\begin{minipage}[c]{0.34\linewidth}
\includegraphics[width=\linewidth]{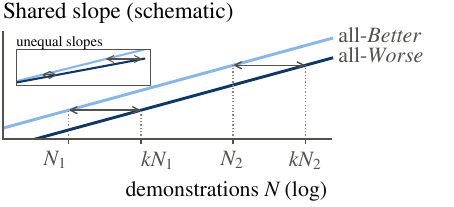}
\end{minipage}\hfill
\begin{minipage}[c]{0.63\linewidth}
\caption{\textbf{Reading the data multiple as a horizontal distance.} Schematic success log-odds against log dataset size. With parallel lines, as in the shared-slope fits, the all-\Worse{} set needs $kN$ demonstrations to match the all-\Better{} success at any size $N$ (arrows). Inset: with unequal slopes, $k$ grows with $N$.}
\label{fig:schematic}
\end{minipage}
\end{figure}

\begin{figure}[t]
  \centering
  \includegraphics[width=\linewidth]{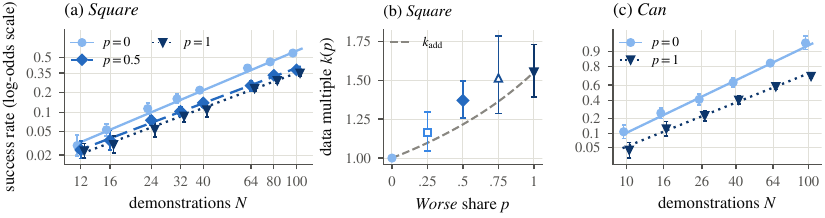}
  \caption{\textbf{Scaling curves and data multiples at 25\,s.}
  (a)~\Square{}: seed mean of the continuity-corrected success log-odds against dataset size for $p=0$, 0.5, and 1, with seed intervals on that scale; the vertical axis is labeled with the corresponding success rates, points are offset along $N$ for legibility, and lines are separate least-squares fits to these means.
  (b)~Data multiple $k(p)$ for the five \Square{} sets, with seed intervals; open markers: the two sets trained on 12 seeds; dashed: additive prediction $\kadd(p)$ computed from the plotted $k(1)$.
  (c)~\Can{}, as in (a), for $p=0$ and 1.}
  \label{fig:scaling}
\end{figure}

\paragraph{The data multiple.}
On \Square{} at 25\,s, the all-\Worse{} multiple $k(1)=1.55$ means that about 62 \Worse{} demonstrations do the work of 40 \Better{} ones.
We read data multiples as horizontal distances between scaling curves (\cref{fig:schematic}).
For each training seed $s$, we fit success log-odds against $\ln N$, with one line per \Worse{} share, a shared slope, and separate intercepts.
A horizontal distance $\ln k_s(p)$ between a set's line and the all-\Better{} line means that, in seed $s$, the set needs $k_s(p)$ times as many demonstrations for the same success rate.
One of its demonstrations is then worth $1/k_s(p)$ all-\Better{} demonstrations.
Across seeds, we report the data multiple $k(p)$ as the geometric mean of $k_s(p)$; $k(1)$ is the all-\Worse{} multiple.
\Cref{fig:scaling}a draws a separate least-squares line through each set's seed-mean log-odds to show the curve shape; the multiples come from the shared-slope fits.
Brackets give 95\% intervals across training seeds, which we call seed intervals.
Unless marked, they are pointwise Student $t$ intervals, paired by seed for changes.

\paragraph{Additive prediction and excess.}
Suppose that every demonstration in a mixed set kept the value it has in a set from its own group.
In seed $s$, the mixed set's multiple would then be the additive prediction $k_{\mathrm{add},s}(p)$, with $1/k_{\mathrm{add},s}(p)=(1-p)+p/k_s(1)$.
The excess is the percentage of extra demonstrations a mixed set needs beyond its additive prediction.
Each seed contributes the log ratio $\ln[k_s(p)/k_{\mathrm{add},s}(p)]$.
Across seeds, the log excess $E$ is the mean of these log ratios, and the excess is $100[\exp(E)-1]\%$.
The same formula applied to the summary multiples gives $\kadd(p)$, with $1/\kadd(p)=(1-p)+p/k(1)$.
For example, $k(1)=1.55$ on \Square{} at 25\,s gives $1/\kadd(0.5)=0.5+0.5/1.55=0.82$, so $\kadd(0.5)=1.22$.
The all-\Better{} and all-\Worse{} sets each pool two teleoperators, so the excess measures what pooling all four costs beyond the two groups' separate values.

\paragraph{Worked example.}
Take one \Square{} seed at 25\,s.
Its all-\Worse{} line lies 0.73 to the right of its all-\Better{} line on the $\ln N$ axis, so $k_s(1)=\exp(0.73)\simeq2.07$ and a \Worse{} demonstration is worth $1/k_s(1)\simeq0.48$ all-\Better{} demonstrations.
The additive prediction for the half-\Better{}, half-\Worse{} set is then $k_{\mathrm{add},s}(0.5)=1/[0.5+0.5/k_s(1)]\simeq1.35$.
Its fitted line gives $k_s(0.5)\simeq1.68$, so its log excess is $E_s(0.5)=\ln[k_s(0.5)/k_{\mathrm{add},s}(0.5)]\simeq0.22$, computed from unrounded values.
The excess of this example seed is thus about 24\%.
The 13\% of \cref{sec:mixing} instead averages the log excess over all seeds first and then converts that mean to a percentage.

\paragraph{Why a shared slope.}
A shared slope summarizes each set by one multiple over the tested sizes, while separate slopes let the multiple change with $N$.
We test this summary in a second evaluation that pools the last five checkpoints of each policy (the five-checkpoint evaluation, \cref{app:five-checkpoint}).
In that evaluation, shared slopes fitted on the smaller sizes predict the success log-odds at the larger sizes more accurately than separate slopes do (\cref{app:slope-checks}).

\paragraph{Slope differences.}
In the final-checkpoint evaluation on \Square{} at 25\,s, the four sets that contain \Worse{} demonstrations have shallower lines than the all-\Better{} set on average.
A multiple that followed this mean slope difference, instead of the shared slope, would grow by 31\% [5\%, 68\%] from $N=12$ to 100 (\cref{app:slope-checks}).
The reported $k(p)$ is therefore a summary over the tested sizes.
\Cref{app:stats} also reads multiples from each set's own success curve and from count models of the success counts.

\section{On \texorpdfstring{\Square{}}{Square}, Most of the Gap Is Late Completion}
\label{sec:speed}

\paragraph{Time limits on \texorpdfstring{\Square{}}{Square}.}
More time removes most of the gap between the two groups on \Square{}, which we measure by $\ln k(1)$ (\cref{fig:horizon}a).
The all-\Worse{} multiple falls from 1.55 [1.39, 1.73] at 25\,s to 1.11 [1.02, 1.22] at 100\,s, removing 75\% [62\%, 94\%] of $\ln k(1)$ (\cref{tab:replay}).
The decline then flattens, and at 200\,s the multiple is 1.08 [0.99, 1.18].
This fall and the excess of \cref{sec:mixing} both exclude zero after a Bonferroni correction across three quantities (\cref{app:multiplicity}).

\begin{figure}[t]
  \centering
  \begin{minipage}[c]{0.56\linewidth}
  \includegraphics[width=\linewidth]{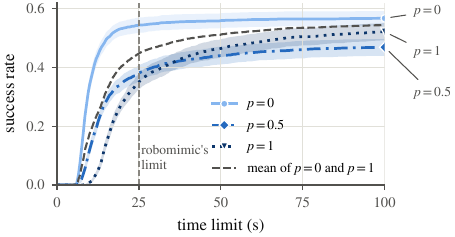}
  \end{minipage}\hfill
  \begin{minipage}[c]{0.41\linewidth}
  \caption{\textbf{Success rate of \Square{} policies at $N=100$ against the time limit.}
  Means over 24 seeds with seed intervals for $p=0$, 0.5, and 1.
  Dashed grey: mean of the $p=0$ and $p=1$ success-rate curves, not the additive prediction.
  Dashed vertical line: robomimic's 25\,s limit.}
  \label{fig:completion}
  \end{minipage}
\end{figure}

\paragraph{Late completion.}
The gap shrinks because policies trained on \Worse{} demonstrations finish later (\cref{fig:completion}).
At $N=100$, the all-\Worse{} success rate rises from 35\% within 25\,s to 52\% within 100\,s, while the all-\Better{} rate rises only from 55\% to 57\%.
Most of the all-\Worse{} deficit thus disappears with the longer limit.
In \cref{fig:completion}, the all-\Better{} curve is nearly flat after robomimic's limit, while the all-\Worse{} curve keeps rising.

\paragraph{Slower execution.}
All-\Worse{} policies are slower even within a single grasp, as the \Worse{} teleoperators are.
One-grasp successes reach the goal in a single sustained grasp.
Counting only one-grasp successes, the time from grasp to success at $N=100$ is 1.95 [1.84, 2.06] times as long for all-\Worse{} as for all-\Better{} policies.
In the demonstrations themselves, the same ratio is 1.74 [1.57, 1.91].
Late completions also include interrupted successes, which reach the goal after a regrasp or a grasp loss (\cref{app:events}).

\paragraph{Diffusion policy.}
A second policy class, a diffusion policy~\citep{chi2025diffusionpolicy}, also shows the decline.
In a separate \Square{} experiment with $N\in\{24,100\}$ and 8 paired seeds, the log of its all-\Worse{} multiple falls by 0.40 [0.31, 0.49] from 25\,s to 100\,s (\cref{app:diffusion}).

\paragraph{\texorpdfstring{\Can{}}{Can}.}
On \Can{}, the same extension removes much less of the gap.
The all-\Worse{} multiple falls from 1.88 [1.72, 2.07] at 25\,s to 1.61 [1.48, 1.77] at 100\,s, removing 24\% [19\%, 30\%] of $\ln k(1)$ (\cref{tab:replay}).
All-\Worse{} \Can{} policies also finish later.
Summed over all sizes and seeds, late completions make up 14\% and 2\% of the all-\Worse{} and all-\Better{} successes within 100\,s (\cref{tab:timing-details}).

\section{The Half-\texorpdfstring{\Better{}}{Better}, Half-\texorpdfstring{\Worse{}}{Worse} Set Needs More Demonstrations Than Its Additive Prediction}
\label{sec:mixing}

\paragraph{The excess.}
Pooling all four teleoperators costs demonstrations beyond the two groups' separate values (\cref{fig:scaling}b, \cref{tab:comparisons-main}).
At 25\,s on \Square{}, the half-\Better{}, half-\Worse{} set needs 13\% [6\%, 21\%] more demonstrations than its additive prediction.
At 25\,s, the 75\%-\Worse{} set also lies above its prediction (\cref{fig:scaling}b, \cref{app:limits}).
Reading each set's size at a target success rate from its own monotone success curve gives excesses of similar size (\cref{tab:comparisons-main}).

\paragraph{Longer time limits.}
The excess persists while the gap between the groups shrinks (\cref{fig:horizon}).
In the final-checkpoint evaluation, at every time limit from 20\,s to 200\,s, it lies between 13\% and 14\%.
Paired by seed, the log excess $E$ changes by 0.00 [\ensuremath{-}0.03, 0.03] from 25\,s to 100\,s.

\paragraph{\texorpdfstring{\Can{}}{Can} and the diffusion policy.}
On \Can{}, the half-\Better{}, half-\Worse{} set needs 9\% [1\%, 17\%] more demonstrations than its additive prediction at both 25\,s and 100\,s (\cref{tab:replay}).
In the diffusion-policy experiment of \cref{sec:speed}, the excess of both the diffusion policy and a matched BC-RNN-GMM has positive point estimates at 25\,s and 100\,s, with seed intervals that include zero (\cref{tab:diffusion}).

\paragraph{Below the all-\Worse{} set.}
At $N=100$ and 100\,s, half-\Better{}, half-\Worse{} policies succeed less often than all-\Worse{} policies (\cref{fig:completion}).
They solve 47\% and 52\% of episodes, a seed-paired difference of \ensuremath{-}5 [\ensuremath{-}9, \ensuremath{-}1] percentage points.
For $k(1)\geq1$, the additive prediction values a demonstration in the half-\Better{}, half-\Worse{} set at $0.5+0.5/k(1)\geq1/k(1)$ all-\Better{} demonstrations, at least the value of a demonstration in the all-\Worse{} set.
Under this prediction, the half-\Better{}, half-\Worse{} set would succeed at least as often as the all-\Worse{} set of the same size.

\paragraph{Training samples.}
Under the default sampler, the longer \Worse{} demonstrations supply 66\% of the training sequences of a half-\Better{}, half-\Worse{} set.
Could the excess only reflect that this sampler draws the \Worse{} demonstrations more often?
First, we keep the policies of the main experiment and change the reference prediction.
The training-sample share prediction credits each group by its share of training sequences instead of its share of demonstrations.
Against it, the excess is 6\% at 25\,s and 11\% at 100\,s, against 13\% and 13\% with the additive prediction (\cref{tab:comparisons-main,tab:sample-shares}).

\begin{figure}[t]
  \centering
  \begin{minipage}[c]{0.5\linewidth}
  \includegraphics[width=\linewidth]{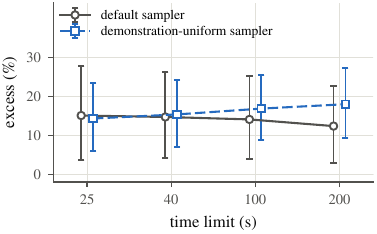}
  \end{minipage}\hfill
  \begin{minipage}[c]{0.46\linewidth}
  \caption{\textbf{Excess of the half-\Better{}, half-\Worse{} set in the sampler experiment on \Square{}.}
  Each sampler's own shared-slope fits on the same 12 seeds, with seed intervals; points are offset along the time axis for legibility (\cref{tab:sampler-fits}).}
  \label{fig:sampler-excess}
  \end{minipage}
\end{figure}

\paragraph{Sampler experiment.}
Second, we retrain with a demonstration-uniform sampler, which draws a demonstration uniformly and then a start position within it.
Each group's share of training sequences then equals its share of demonstrations, so the two reference predictions coincide.
In this separate \Square{} experiment, we train the all-\Better{} and all-\Worse{} sets and the half-\Better{}, half-\Worse{} set at all eight sizes with each sampler, 576 policies on 12 seeds.
The two policies of each seed, set, and size differ only in the sampler, and each sampler has its own shared-slope fits (\cref{app:sampling}).
The answer is no: the excess is present under both samplers (\cref{fig:sampler-excess}).

The excess at 25\,s is 14\% [6\%, 23\%] with the demonstration-uniform sampler and 15\% [4\%, 28\%] with the default sampler in this experiment (\cref{tab:sampler-fits}).
Under both samplers, its interval excludes zero at all four time limits.
Paired by seed, the ratio of required to predicted demonstrations, which is one plus the excess, changes by \ensuremath{-}1\% [\ensuremath{-}12\%, 12\%] at 25\,s between the samplers.
This is a relative change in that ratio, not a difference in percentage points.

\paragraph{Pooling within a group.}
At $N=40$, pooling also lowers success within the \Better{} group.
On 12 paired seeds, single-teleoperator \Better{} policies have success log-odds 0.60 [0.20, 0.99] higher at 25\,s than the policy trained on B1 and B2 together (\cref{app:teleoperators}).
The excess also differs between the four teleoperator pairs that combine one \Better{} and one \Worse{} teleoperator (\cref{app:teleoperators}).

\begin{table}[t]
  \centering\small
  \setlength{\tabcolsep}{4pt}
  \begin{tabular}{@{}>{\raggedright\arraybackslash}p{0.47\linewidth}ccc@{}}
  \toprule
  Estimate & 25\,s & 100\,s & 200\,s \\
  \midrule
  Shared slope, additive prediction & 13 [6, 21]& 13 [7, 20]& 14 [8, 20]\\
  Shared slope, training-sample share prediction & 6 [\ensuremath{-}1, 13]& 11 [5, 18]& -- \\
  \Can{}: shared slope, additive prediction & 9 [1, 17]& 9 [1, 17]& 10 [2, 18]\\
  \midrule
  Monotone-curve inversion, $u=0.10$ & 15 [2, 28]& -- & -- \\
  Monotone-curve inversion, $u=0.20$ & 13 [5, 25]& -- & -- \\
  Monotone-curve inversion, $u=0.30$ & 16 [8, 27]& 18 [8, 30]& 16 [8, 29]\\
  \bottomrule
  \end{tabular}
  \caption{\textbf{Excess (\%) of the half-\Better{}, half-\Worse{} set under the default sampler.}
  Columns: time limits. Rows are \Square{} unless marked. The first two rows apply two reference predictions to the same per-seed shared-slope fits; the inversion rows read each set's size at target success rate $u$ from its own monotone success curve.
  Brackets: seed intervals; percentiles of resampled seeds for the \Can{} and inversion rows. Dashes: the training-sample share prediction is computed through 100\,s; inversions at other targets and limits are in \cref{tab:equal-success-mixing}.}
  \label{tab:comparisons-main}
\end{table}
\section{What the Multiples Mean for Evaluation and Collection}
\label{sec:implications}

\paragraph{Evaluation.}
Comparisons of data sources should report success across time limits, or throughput.
Throughput counts successes per unit of execution time, with each failed episode counted at the full limit.
Late completion still costs execution time: at $N=100$ and a 100\,s limit, all-\Worse{} \Square{} policies achieve 26\% [15\%, 35\%] fewer successes per unit of execution time than all-\Better{} policies (\cref{app:throughput}).
Scoring only at robomimic's 25\,s limit counts late completions as failures, yet on \Square{} late completion accounts for most of the gap in $\ln k(1)$ between the groups (\cref{sec:speed}).

\paragraph{Collection.}
\Worse{} demonstrations that a team has already recorded raised success when added to a \Better{} training set.
Each half-\Better{}, half-\Worse{} training set of size $2n$ contains the all-\Better{} set of size $n$, so its policy shows the effect of adding $n$ \Worse{} demonstrations to that set.
For the four such additions ($12\to24$, $16\to32$, $32\to64$, $40\to80$), the seed-mean gain in success log-odds is positive at 25, 40, and 100\,s, and every interval excludes zero (\cref{app:additions}).
\Cref{box:takeaways} converts the multiples into break-even pay ratios, which depend on the decision and the time limit.

\begin{figure}[t]
  \centering
  \fbox{\begin{minipage}{0.96\linewidth}\small
  \refstepcounter{infobox}\label{box:takeaways}%
  \textbf{Box~\theinfobox: When is recording \Worse{} demonstrations cheaper on \Square{}?}\par\smallskip
  The break-even pay ratio is the \Worse{}/\Better{} hourly pay ratio below which recording \Worse{} demonstrations costs less for the same policy success.
  A gain of $x$ \Better{}-equivalent demonstrations means that, under the shared slope, adding $x$ all-\Better{} demonstrations would reach the same success rate.
  \emph{Replace:} build a new dataset from \Worse{} instead of \Better{} demonstrations, paying for every demonstration.
  \emph{Add:} keep an existing \Better{} set and pay only for an equal number of new \Worse{} demonstrations.\par\smallskip
  With recording time as the only cost, the break-even pay ratio is $1/(rm)$, where $r=1.93$ is the \Worse{}/\Better{} ratio of mean demonstration durations and $m$ is the number of \Worse{} demonstrations per \Better{}-equivalent demonstration.
  \emph{Replace} at 25\,s: one \Better{} demonstration is worth $m=k(1)=1.55$ \Worse{} demonstrations, which take $1.55\times1.93\approx3$ times the recording time, so the ratio is about a third.\par
  \emph{Add} at 25\,s: the existing 40 \Better{} demonstrations are already paid for.
  The 40 new \Worse{} demonstrations add 18.4 [13.5, 23.7] \Better{}-equivalent demonstrations to them, so $m=40/18.4\approx2.2$ and the ratio is about 0.24; at 100\,s, $m\approx1.5$ (\cref{app:costs}).
  Adding is stricter because the excess makes each added \Worse{} demonstration worth less than a demonstration in an all-\Worse{} set.\par\smallskip
  {\centering
  \setlength{\tabcolsep}{5pt}
  \begin{tabular}{@{}>{\raggedright\arraybackslash}p{0.47\linewidth}cc@{}}
    \toprule
    Decision & 25\,s limit & 100\,s limit \\
    \midrule
    Replace & 0.33 [0.29, 0.38]& 0.47 [0.41, 0.53]\\
    Add & 0.24 [0.18, 0.32]& 0.35 [0.28, 0.44]\\
    \bottomrule
  \end{tabular}\par}\smallskip
  Both ratios are below one because \Worse{} demonstrations take longer to record and are worth less.
  The ratios use the default sampler and sizes $N=12$--100, and their intervals include variation in demonstration duration.
  \Cref{app:costs} adds collection overheads and throughput.
  \end{minipage}}
\end{figure}
\section{Related Work}
\label{sec:related}

\paragraph{Data value in units of data.}
\Citet{hernandez2021scaling} define the effective data transferred by pre-training as the extra fine-tuning data that a model of the same size, trained from scratch, would need to reach the same loss.
They find that pre-training effectively multiplies the fine-tuning set in the low-data regime.
\Citet{lin2024data} study data scaling laws of imitation learning for robotic manipulation and leave the effect of data quality on these laws to future work.
We apply the effective-data accounting to closed-loop success for data from teleoperators of different experience, where the multiple depends on the time limit and on how the groups are pooled.

\paragraph{Teleoperator quality.}
Robomimic reports that its lower-quality datasets contain demonstrations that take more time~\citep{mandlekar2021matters}.
Its low-dimensional BC-RNN succeeds more often on \Better{} than on \Worse{} data and improves when \Worse{} demonstrations are added to \Better{} ones.
\Citet{gandhi2022eliciting} show that adding demonstrations to a base set can raise or lower success depending on which teleoperator recorded them, and use a compatibility measure with the base policy to filter demonstrations and guide collection.
\Citet{belkhale2023data} relate data quality to action consistency and, on robomimic's labeled datasets, find that demonstration duration increases from \Better{} to \Worse{}.
We compare sets of fixed total size with the additive prediction and report additions to \Better{} sets separately (\cref{sec:implications}).

\paragraph{Curation and weighting.}
DemInf scores each demonstration by its contribution to the mutual information between states and actions, and checks the scores against robomimic's teleoperator labels~\citep{hejna2025robot}.
CUPID scores demonstrations by their estimated influence on closed-loop performance~\citep{agia2025cupid}, and other methods filter demonstrations by smoothness~\citep{kulkarni2026learning}, estimate each teleoperator's expertise jointly with the policy~\citep{beliaev2022imitation}, or learn confidence weights over state--action pairs~\citep{zhang2021confidence}.
ReMix learns domain weights for a pre-training mixture and notes that sampling state--action pairs uniformly weights robomimic's teleoperators unevenly, because some take longer to complete the task~\citep{hejna2025remix}.
With the data fixed, our sampler experiment gives every demonstration the same weight, and the excess of the half-\Better{}, half-\Worse{} set remains (\cref{sec:mixing}).

\paragraph{Execution speed.}
SAIL and DemoSpeedup train policies to run faster than their demonstrations~\citep{ranawaka2025sail,guo2025demospeedup}, and \citet{shi2025diversity} resample demonstration chunks in time during training, using an action rate predicted from the observation.
Our time limits instead vary how long a fixed policy may run.

\paragraph{Diversity among demonstrations.}
\Citet{luo2026geometric} vary the shape diversity of trajectories at fixed dataset sizes, using motion-planner demonstrations in simulation and demonstrations from three people on a real arm.
Success rates often rise and then fall as the geometric entropy of the trajectories grows, a fall they attribute to demonstrations that hold several strategies for one conditioning signal.
This entropy is computed on trajectories resampled by arc length, so it leaves out the speed difference in our mixed sets, whose \Worse{} demonstrations are slower than their \Better{} ones (\cref{sec:speed}).
\Citet{shi2025diversity} introduce the resampling described above to counter differences in action rate among teleoperators, which give identical inputs different action chunks and destabilize training.
Pooling all four of our teleoperators costs demonstrations beyond the additive prediction (\cref{sec:mixing}), consistent with the explanation of \citet{shi2025diversity} based on differences in action rate.
\section{Limitations and Conclusion}
\label{sec:limitations}

\paragraph{Limitations.}
The measurements cover two of robomimic's simulated multi-human tasks, low-dimensional observations, and the four teleoperators B1, B2, W1, and W2.
The policies are BC-RNN-GMM trained with robomimic's recipe under two samplers, and a diffusion policy repeats the \Square{} time-limit result.
The seed intervals describe variation across training seeds for these fixed demonstrations and teleoperators.
The break-even pay ratios are computed from the measured multiples and the recorded demonstration durations.

\paragraph{Conclusion.}
On \Square{}, most of the gap between \Better{} and \Worse{} demonstrations at robomimic's 25\,s limit is late completion, which a longer limit counts as success; on \Can{}, most of the gap remains.
Pooling all four teleoperators in a half-\Better{}, half-\Worse{} set costs an excess that persists from 20\,s to 200\,s under the default sampler and remains under the demonstration-uniform sampler.
Teams that collect demonstrations can report success at several time limits or as throughput (\cref{sec:speed,sec:implications}) and check a mixed set against its additive prediction (\cref{sec:mixing}).
They can price \Worse{} teleoperators with the break-even pay ratio of the decision at hand (\cref{box:takeaways}).

\clearpage

\bibliography{main}

\clearpage
\appendix
\counterwithin{table}{section}
\counterwithin{figure}{section}
\renewcommand{\thetable}{\thesection\arabic{table}}
\renewcommand{\thefigure}{\thesection\arabic{figure}}
\setlength{\intextsep}{6pt plus 2pt minus 2pt}
\setlength{\textfloatsep}{8pt plus 2pt minus 2pt}
\setlength{\floatsep}{6pt plus 2pt minus 2pt}
\renewcommand{\arraystretch}{0.9}
\AddToHook{env/table/begin}{%
  \renewcommand{\small}{\fontsize{8}{9.5}\selectfont}%
  \renewcommand{\footnotesize}{\fontsize{8}{9.5}\selectfont}%
}
\AddToHook{env/tabular/before}{\setlength{\belowcaptionskip}{3pt}}
\section{Terms and Notation}
\label{app:terms}

\Cref{tab:terms} collects the terms and symbols of the main text.

\begin{table}[!ht]
\centering\small
\setlength{\tabcolsep}{3pt}
\begin{tabular}{@{}>{\raggedright\arraybackslash}p{0.25\linewidth}>{\raggedright\arraybackslash}p{0.71\linewidth}@{}}
\toprule
Term & Definition \\
\midrule
Dataset size $N$ & Number of demonstrations in a training set. \\
\Worse{} share $p$ & Fraction of a training set's demonstrations recorded by the \Worse{} teleoperators: $p=0$ for the all-\Better{} set, $p=0.5$ for the half-\Better{}, half-\Worse{} set, and $p=1$ for the all-\Worse{} set. \\
Time limit & Longest execution time within which a success counts: 15, 20, 25, 40, 100, and 200\,s are 300, 400, 500, 800, 2000, and 4000 control steps at 20\,Hz. \\
Late completion & A success after 25\,s and within a longer time limit. \\
Data multiple $k(p)$ & Factor by which a set with \Worse{} share $p$ must be larger than the all-\Better{} set for its policies to reach the same success rate, read from the shared-slope fit of each seed as $k_s(p)$; $k(p)$ is the geometric mean over seeds. \\
All-\Worse{} multiple $k(1)$ & The data multiple of the all-\Worse{} set. \\
Additive prediction $\kadd(p)$ & The data multiple of a mixed set if each demonstration kept the worth, in all-\Better{} demonstrations, that it has in a set of its own group: $1/\kadd(p)=(1-p)+p/k(1)$. \\
Log excess $E(p)$ & Mean over seeds of $\ln[k_s(p)/k_{\mathrm{add},s}(p)]$. \\
Excess & Extra demonstrations a mixed set needs beyond its additive prediction, in percent: $100[\exp(E)-1]\%$. \\
Training-sample share prediction & A reference prediction like $\kadd(p)$ that weights each group by its share $q$ of training sequences instead of its share $p$ of demonstrations. \\
Monotone-curve inversion & Ratio of dataset sizes at target success rate $u$, read from each set's own monotone success curve. \\
Default sampler & Robomimic's sampler: training sequences drawn uniformly over all start positions, so a demonstration is drawn in proportion to its length. \\
Demonstration-uniform sampler & A demonstration drawn uniformly, then a start position within it. \\
Break-even pay ratio & The \Worse{}/\Better{} hourly pay ratio below which recording \Worse{} demonstrations costs less at equal success. \\
Seed interval & 95\% interval across training seeds; Student $t$, paired by seed for changes, unless a table marks percentile intervals from resampled seeds. \\
\bottomrule
\end{tabular}
\caption{\textbf{Terms and symbols.}}
\label{tab:terms}
\end{table}

\paragraph{Per-seed and summary values.}
The reported excess at 25\,s, 13\%, converts the seed mean of the log excess $E_s(0.5)$ to a percentage; \cref{sec:setup} works through one seed.
The ratio of the summary values, $k(0.5)/\kadd(0.5)$ with $\kadd(0.5)$ computed from the summary $k(1)$, exceeds one by 13\% at 25\,s.
The two agree to the nearest percent here, but they can differ because $\kadd$ is nonlinear in $k(1)$ and each seed's prediction uses its own $k_s(1)$.
For the same reason, the vertical distance between a marker and the dashed curve in \cref{fig:scaling}b differs slightly from the excess.
\section{Experimental Details}
\label{app:details}

\paragraph{Data and software.}
We train on the low-dimensional observations of robomimic's v0.2 multi-human datasets.
Because we simulate with robosuite 1.5.1, we recompute the object observations at each evaluation step and arrange them in the v0.2 layout.
On \Square{}, robomimic's v1.5 datasets, whose observations are extracted again for robosuite~1.5, have the same demonstrations, actions, and initial states as v0.2.
Software: robomimic 0.5.0, robosuite 1.5.1~\citep{zhu2020robosuite}, MuJoCo 3.2.3~\citep{todorov2012mujoco}, and PyTorch 2.4.1.
On \Square{}, the \Worse{}/\Better{} ratio of mean demonstration lengths is 1.93.
13/100 \Square{} \Worse{} demonstrations exceed 25\,s, against 0/100 \Square{} \Better{} and 10/100 \Can{} \Worse{} demonstrations.
Pauses explain little of the length difference.
Removing the steps with stored end-effector speed below 0.02\,m/s from both groups lowers the \Square{} mean-length ratio only to $1.810\,[1.665, 1.971]$, and the \Can{} \Worse{}/\Better{} ratio after removal is $1.977\,[1.808, 2.169]$.
\space
These lengths count complete recordings, including the steps after success.
Brackets are percentile intervals from 20{,}000 draws that resample demonstrations within each teleoperator.

\paragraph{Training sets.}
Training sets draw from every released demonstration, including the validation split.
At $N=100$, the all-\Better{} and all-\Worse{} sets use all demonstrations of their group, and mixed sets draw subsets, so their seeds vary both the demonstrations and the training randomness.
Each seed randomly orders the demonstrations within each teleoperator and alternates the two teleoperators of a group to fill its quota.
When a group's quota is odd, the parity of the seed decides which teleoperator contributes the extra demonstration.
Sets are nested in $N$ (\cref{tab:protocols}).

\paragraph{Training and evaluation.}
The BC-RNN-GMM recipe~\citep{mandlekar2021matters} uses a 2-layer, 400-unit LSTM, a 5-mode Gaussian-mixture head, 10-step sequences, and Adam at $10^{-4}$.
Training runs for 2000 epochs of 100 gradient steps~\citep{mandlekar2021matters}, and we use batches of 100.
The unidirectional LSTM feeds the GMM head directly.
The GMM uses softplus standard deviations with floor $10^{-4}$.
Adam uses a constant learning rate, zero weight decay, and no gradient clipping.
Observation and action normalization are disabled; sequences repeat boundary frames when padding.
\space
As in robomimic, evaluation uses low-noise actions, and the policy outputs actions at 20\,Hz~\citep{mandlekar2021matters}.
We reset the recurrent state every 10 steps and record the step of first success.
Compared policies are evaluated from the same initial states.

\paragraph{Training devices.}
The main experiment is trained on two GPU machines and one CPU machine.
Within each task, all sets and sizes of a seed are trained on one machine, except for one \Square{} seed.
For this seed, the all-\Better{} and all-\Worse{} sets are trained on a GPU machine and the half-\Better{}, half-\Worse{} set on the CPU machine, so the \Worse{} share and the training machine change together.
\space
In the single-teleoperator and teleoperator-pair experiments, all compared policies are trained on the same machine and GPU model (\cref{app:teleoperators}).
The sampler experiment trains and evaluates all policies of a seed on one CPU model (\cref{app:sampling}).

\begin{table}[!ht]
\centering\small
\setlength{\tabcolsep}{3pt}
\begin{tabular}{@{}lp{0.74\linewidth}@{}}
\toprule
Design & Values \\
\midrule
\Square{} & $p\in\{0, 0.25, 0.5, 0.75, 1\}$, $N\in\{12, 16, 24, 32, 40, 64, 80, 100\}$; 24 seeds for $p\in\{0, 0.5, 1\}$ and the first 12 for the other two shares; 768 policies. \\
\Can{} & $p\in\{0, 0.5, 1\}$, $N\in\{10, 16, 26, 40, 64, 100\}$; 12 seeds; 216 policies. \\
\midrule
Evaluation & Checkpoints, episodes per policy, time limits \\
\midrule
Final checkpoint & Epoch 2000; 100 initial states, a subset of the five-checkpoint states; up to 200\,s. \\
Five checkpoints & 5 checkpoints at epochs 1600--2000; 500 episodes; 15--40\,s. \\
\bottomrule
\end{tabular}
\caption{\textbf{Design and evaluations of the main experiment.} The final-checkpoint evaluation gives the main-text results unless marked; the five-checkpoint evaluation repeats the analysis in \cref{app:five-checkpoint}.}
\label{tab:protocols}
\end{table}

\begin{table}[!ht]
\centering\small
\setlength{\tabcolsep}{3pt}
\begin{tabular}{@{}lcccc@{}}
\toprule
Experiment & Seeds & Policies & \shortstack{Five-ckpt.\\episodes} & \shortstack{Final-ckpt.\\episodes} \\
\midrule
Six teleoperators, $N=40$ & 3 & 18 & 500 & 100 \\
Teleoperator pairs, $N=64,100$ & 6 & 48 & 500 & 100 \\
Sampler experiment, both samplers, eight sizes & 12& 576& -- & 100\\
Single teleoperators and groups, $N=40$ & 12 & 72 & 500 & 100 \\
Teleoperator pairs and groups, $N=100$ & 12 & 72 & 500 & 100 \\
Single teleoperators and groups, three sizes & 6 & 108 & 500 & -- \\
Teleoperator pairs, four sizes & 4 & 64 & 500 & -- \\
Diffusion and paired BC-RNN-GMM, two sizes & 8 & 48 / 48 & -- & 100 \\
4000-epoch training & 24 / 12 & 768 & 250 & -- \\
Exposure-matched all-\Worse{}, two sizes & 8 & 16 & 500 & 100 \\
\bottomrule
\end{tabular}
\space
\caption{\textbf{Experiments beyond the main experiment.} Seeds, trained policies, and evaluation episodes per policy; all on \Square{}. Sampler experiment: \cref{app:sampling}; longer training and exposure matching: \cref{app:exposure}.}
\label{tab:inventory}
\end{table}

\subsection{Comparison with robomimic's published results}
\label{app:known}
Under robomimic's training split and best-checkpoint protocol, our success rates are within 0.06 of its published values on \Square{} and within 0.07 on \Can{} (\cref{tab:known}).
The protocol evaluates 50 episodes every 50 epochs, with a 25\,s limit for the multi-human (MH) data and 20\,s for the proficient-human (PH) data~\citep{mandlekar2021matters}.
On \Can{}, both intervals over 3 seeds include the published values.

\begin{table}[!ht]
  \centering\small
\begin{tabular}{@{}lcccc@{}}
\toprule
Dataset & Demonstrations & Seeds & Ours & Published \\
\midrule
\Square{} MH \Better{} & 90 & 3 & $0.613\,[0.537, 0.689]$ & 0.660 \\
\Square{} MH \Worse{} & 90 & 3 & $0.420\,[0.289, 0.551]$ & 0.393 \\
\Square{} PH 20\% & 36 & 3 & $0.333\,[0.230, 0.437]$ & 0.387 \\
\Square{} PH 50\% & 90 & 1 & $0.700$ & 0.673 \\
\Square{} PH 100\% & 180 & 1 & $0.880$ & 0.840 \\
\midrule
\Can{} MH \Better{} & 90& 3& 0.987 [0.958, 1.015]& $0.993\pm0.009$ \\
\Can{} MH \Worse{} & 90& 3& 0.853 [0.653, 1.054]& $0.920\pm0.016$ \\
\bottomrule
\end{tabular}
\space
  \caption{\textbf{Our training against robomimic's published success rates on \Square{} and \Can{}.} Maximum success rate over training under robomimic's protocol, mean over seeds, against the published BC-RNN values~\citep{mandlekar2021matters}. Success rates are fractions. PH 20\%, 50\%, and 100\%: robomimic's subsets of the proficient-human data. Brackets: $t$ intervals across the seeds' selected maxima, not clipped to $[0,1]$. $\pm$ as reported in~\citep{mandlekar2021matters}.}
  \label{tab:known}
\end{table}
\section{Estimation and Robustness}
\label{app:stats}

\subsection{Per-seed fits}
\label{app:per-seed}
For each seed $s$ and \Worse{} share $p$, we fit the line $c_{sp}+\beta_{sp}\ln N$ across sizes $N$ to the corrected log-odds $z_{spN}=\ln[(y+0.5)/(R-y+0.5)]$.
Here $y$ is the success count out of $R$ episodes; adding 0.5 to both the success and the failure count keeps the log-odds finite.
The shared slope $\bar\beta_s^{(\mathrm{all})}$ fits all sets of the seed with separate intercepts, and $\bar\beta_s^{(3)}$ uses only $p\in\{0, 0.5, 1\}$.
For mean log-odds $\bar z_{sp}$ over sizes, $\ln k_s(p)=(\bar z_{s0}-\bar z_{sp})/\bar\beta_s^{(\mathrm{all})}$, and we exponentiate the seed mean of $\ln k_s(p)$ to obtain $k(p)$.
The log excess is $E(p)=\operatorname{mean}_s E_s(p)$ with $E_s(p)=\ln k_s(p)-\ln k_{\mathrm{add},s}(p)$.
Unless marked, multiples, log excesses, and break-even pay ratios use all sets.

With separate lines, the multiple of seed $s$ at all-\Better{} size $N_g$ is
\[
\ln k_{s}^{\mathrm{line}}(p)=\frac{c_{s0}-c_{sp}+(\beta_{s0}-\beta_{sp})\ln N_g}{\beta_{sp}}.
\]
At the geometric-mean size, separate lines give nearly the same multiples as the shared slope.
In the five-checkpoint evaluation at 25\,s, the all-\Worse{} values at this size ($N_g=36$ on \Square{}, 32 on \Can{}) are 1.58 [1.49, 1.68] and 2.04 [1.87, 2.23].
Their ratios to the shared-slope multiples are 0.99 [0.96, 1.02] and 1.05 [0.98, 1.12], and the other \Square{} sets differ by at most 3\%.

\subsection{Shared slope and growth of the multiple}
\label{app:slope-checks}
The shared slope is an approximation: the sets that contain \Worse{} demonstrations have shallower lines on average (\cref{tab:slope-tests}).
For all-\Better{} slope $\beta_0$ and mean slope difference $\Delta\beta$ of the four sets with \Worse{} demonstrations, a $c$-fold increase in all-\Better{} size raises the multiple implied by $\Delta\beta$ by $100[c^{\beta_0/(\beta_0+\Delta\beta)-1}-1]\%$ (\cref{fig:schematic}).
We refit both quantities jointly in the seed-and-set fixed-effect model on 20{,}000 whole-seed draws (\cref{tab:joint-growth}).
The draws are stratified by set coverage, that is, by whether a seed has all five sets or only $p\in\{0, 0.5, 1\}$.
At 200\,s the interval of the growth includes zero.
For held-out prediction, each of the first 12 seeds, which have all five sets, fits $N\leq40$ and predicts the log-odds at $N\geq64$.
Shared slopes give the lower held-out root mean squared error (RMSE), and separate slopes the lower in-sample Akaike information criterion (AIC).

The five-checkpoint mean slope difference comes from fitting seed and set fixed effects to 96 per-seed slopes.
The likelihood comparison fits correlated Gaussian random seed intercepts and slopes by maximum likelihood.
Its null removes the mean slope differences between sets, and we refit both models on 2{,}000 whole-seed draws stratified by set coverage.
A block is one checkpoint together with its evaluation initial states.
Slope sensitivity resamples seeds, blocks, or both in 20{,}000 percentile draws, and each block draw is shared across seeds, sets, and sizes.
The binomial mixed model fits integer counts with block effects and correlated random seed intercepts and slopes by Laplace approximation.
Its Wald interval conditions on the blocks without a dispersion adjustment.
The Pearson dispersion is 1.655 with a shared slope and 1.631 with separate slopes.

\begin{table}[!ht]
\centering\fontsize{8}{9.5}\selectfont
\setlength{\tabcolsep}{3pt}
\begin{tabular}{@{}p{0.56\linewidth}l@{}}
\toprule
Five-checkpoint evaluation, 25\,s & Estimate \\
\midrule
Prediction RMSE, shared / separate & 0.34 / 0.58\\
Paired RMSE difference; seeds favoring shared & \ensuremath{-}0.24 [\ensuremath{-}0.37, \ensuremath{-}0.11]; 10/12\\
AIC decrease, shared$\to$separate, 20/25/40\,s & 7.0 / 4.5 / 5.7\\
Likelihood-ratio statistic; bootstrap $P$ & 12.5; 0.12\\
\bottomrule
\end{tabular}\par\smallskip
\begin{tabular}{@{}lccc@{}}
\toprule
Slope difference & 2000 epochs & 4000 epochs & Paired change \\
\midrule
Mean of four sets & \ensuremath{-}0.14 [\ensuremath{-}0.25, \ensuremath{-}0.03]& \ensuremath{-}0.13 [\ensuremath{-}0.21, \ensuremath{-}0.03] & 0.01 [\ensuremath{-}0.05, 0.08] \\
$p=0.25$ & \ensuremath{-}0.13 [\ensuremath{-}0.31, 0.05]& \ensuremath{-}0.14 [\ensuremath{-}0.26, \ensuremath{-}0.03] & \ensuremath{-}0.01 [\ensuremath{-}0.08, 0.06] \\
$p=0.5$ & \ensuremath{-}0.11 [\ensuremath{-}0.23, 0.01]& \ensuremath{-}0.12 [\ensuremath{-}0.23, \ensuremath{-}0.01] & \ensuremath{-}0.01 [\ensuremath{-}0.11, 0.07] \\
$p=0.75$ & \ensuremath{-}0.16 [\ensuremath{-}0.36, 0.03]& \ensuremath{-}0.18 [\ensuremath{-}0.32, \ensuremath{-}0.03] & \ensuremath{-}0.02 [\ensuremath{-}0.10, 0.08] \\
$p=1$ & \ensuremath{-}0.09 [\ensuremath{-}0.23, 0.05]& \ensuremath{-}0.06 [\ensuremath{-}0.17, 0.07] & 0.03 [\ensuremath{-}0.07, 0.15] \\
\bottomrule
\end{tabular}\par\smallskip
\begin{tabular}{@{}lcccc@{}}
\toprule
Slope sensitivity & Seeds & Blocks & Seeds and blocks & Binomial mixed model \\
\midrule
Mean difference & \ensuremath{-}0.14 [\ensuremath{-}0.23, \ensuremath{-}0.05]& \ensuremath{-}0.14 [\ensuremath{-}0.22, \ensuremath{-}0.04]& \ensuremath{-}0.14 [\ensuremath{-}0.27, 0.01]& \ensuremath{-}0.14 [\ensuremath{-}0.18, \ensuremath{-}0.11] \\
\bottomrule
\end{tabular}
\space
\caption{\textbf{\Square{} slope differences and held-out prediction.} 25\,s unless marked. RMSE differences use seed intervals; 4000-epoch training (\cref{app:exposure}) and slope sensitivity use percentile intervals; the binomial mixed model uses a Wald interval.}
\label{tab:slope-tests}
\end{table}

\begin{table}[!ht]
\centering\small
\setlength{\tabcolsep}{4pt}
\begin{tabular}{@{}llcc@{}}
\toprule
Evaluation & Limit (s) & Growth, \% [joint interval] & Fixed $\beta_0$ interval \\
\midrule
Final checkpoint & 25 & 31 $[5, 68]$ & $[4, 76]$ \\
Final checkpoint & 200 & 22 $[\ensuremath{-}1, 51]$ & $[\ensuremath{-}1, 56]$ \\
Five checkpoints & 25 & 24 $[7, 45]$ & $[7, 46]$ \\
\bottomrule
\end{tabular}
\space
\caption{\textbf{Growth of the \Square{} multiple implied by the mean slope difference of the four sets with \Worse{} demonstrations.} Final-checkpoint rows give the growth from $N=12$ to 100; the five-checkpoint row gives the growth over a tenfold increase in size. Joint and fixed-$\beta_0$ intervals use the same stratified whole-seed draws.}
\label{tab:joint-growth}
\end{table}

\subsection{Observation models}
\label{app:observation}
We compare continuity-corrected least squares (OLS) with binomial and beta-binomial maximum likelihood on the same integer success counts (\cref{tab:observation-models}).
The count models fit the integer successes without continuity correction.
Each model fits separate intercepts and a shared slope within each seed, over every set and size, and multiples and log excesses are computed within seeds before averaging.
On \Square{}, the log excess of the half-\Better{}, half-\Worse{} set has positive seed intervals under all three models at the four time limits of \cref{tab:observation-models}.

\begin{table}[!ht]
\centering\fontsize{8}{9.5}\selectfont
\setlength{\tabcolsep}{4pt}
\begin{tabular}{@{}lccc@{}}
\toprule
Limit (s) & Model & $k(1)$ & $E(0.5)$ \\
\midrule
25 & OLS & 1.55 [1.39, 1.73] & 0.13 [0.06, 0.19] \\
 & Bin. & 1.63 [1.52, 1.74] & 0.16 [0.10, 0.21] \\
 & BB & 1.62 [1.51, 1.74] & 0.16 [0.10, 0.21] \\
40 & OLS & 1.30 [1.18, 1.43] & 0.13 [0.07, 0.19] \\
 & Bin. & 1.35 [1.27, 1.44] & 0.16 [0.10, 0.21] \\
 & BB & 1.35 [1.26, 1.44] & 0.16 [0.10, 0.21] \\
100 & OLS & 1.11 [1.02, 1.22] & 0.13 [0.07, 0.19] \\
 & Bin. & 1.15 [1.08, 1.23] & 0.16 [0.10, 0.21] \\
 & BB & 1.15 [1.07, 1.23] & 0.15 [0.10, 0.21] \\
200 & OLS & 1.08 [0.99, 1.18] & 0.13 [0.07, 0.19] \\
 & Bin. & 1.10 [1.03, 1.18] & 0.15 [0.10, 0.20] \\
 & BB & 1.10 [1.03, 1.18] & 0.15 [0.10, 0.20] \\
\bottomrule
\end{tabular}
\par\smallskip
\begin{tabular}{@{}lccc@{}}
\toprule
Limit (s) & Model & $\Delta\ln k(1)$, model $-$ OLS & $\Delta E(0.5)$, model $-$ OLS \\
\midrule
25 & Bin. & 0.05 [\ensuremath{-}0.01, 0.11] & 0.03 [\ensuremath{-}0.01, 0.06] \\
 & BB & 0.05 [\ensuremath{-}0.01, 0.11] & 0.03 [\ensuremath{-}0.01, 0.06] \\
40 & Bin. & 0.04 [\ensuremath{-}0.02, 0.10] & 0.03 [\ensuremath{-}0.01, 0.06] \\
 & BB & 0.04 [\ensuremath{-}0.02, 0.10] & 0.03 [\ensuremath{-}0.01, 0.06] \\
100 & Bin. & 0.03 [\ensuremath{-}0.02, 0.09] & 0.03 [\ensuremath{-}0.003, 0.06] \\
 & BB & 0.03 [\ensuremath{-}0.02, 0.09] & 0.03 [\ensuremath{-}0.01, 0.06] \\
200 & Bin. & 0.02 [\ensuremath{-}0.03, 0.07] & 0.02 [\ensuremath{-}0.01, 0.06] \\
 & BB & 0.02 [\ensuremath{-}0.03, 0.07] & 0.02 [\ensuremath{-}0.01, 0.05] \\
\bottomrule
\end{tabular}
\space
\caption{\textbf{Observation models on \Square{}.} Bin.: binomial; BB: beta-binomial. All sizes and sets, 24 seeds. Top: all-\Worse{} multiples and log excesses $E(0.5)$; bottom: within-seed changes from OLS on the log scale. Brackets: seed intervals; multiples exponentiate the intervals for mean $\ln k$.}
\label{tab:observation-models}
\end{table}

\subsection{Inverting each set's own success curve}
\label{app:equal-success}
Within the sets' common observed range of seed-mean success, we invert each curve for $k(p,u)=N_p(u)/N_0(u)$ at target success rate $u$, requiring both sizes to lie within the tested sizes of the task.
The monotone-curve inversion uses equal-weight isotonic regression and linear interpolation in $\ln N$; a target equal to a plateau takes its smallest size, and a target above a plateau interpolates from its last point.
The separate-line inversion fits each set's mean corrected log-odds against $\ln N$.
For the all-\Worse{} set, the two inversions give close point estimates on \Square{} and at the outer \Can{} targets, and they differ at the middle \Can{} target (\cref{tab:equal-success-targets}).

For the half-\Better{}, half-\Worse{} set on \Square{}, $k(p,u)$ is the monotone-curve multiple of each seed-mean success curve.
The log excess at a common target is $E^{\mathrm{mono}}(u)=\ln k(0.5,u)+\ln[0.5+0.5/k(1,u)]$, and \cref{tab:equal-success-mixing} reports the excess $100[\exp(E^{\mathrm{mono}}(u))-1]\%$.
$E^{\mathrm{mono}}(u)$ inverts seed-mean curves, whereas $E(p)$ averages per-seed log excesses.

\begin{table}[!ht]
\centering\small
\setlength{\tabcolsep}{3pt}
\begin{tabular}{@{}lccccc@{}}
\toprule
Task, evaluation & $u$ & Monotone-curve $k(1,u)$ & No solution & Separate-line $k(1,u)$ & No solution\\
\midrule
\Square{}, five checkpoints, 25\,s & 0.05 & 1.516 [1.332, 1.708] & 0\% & 1.523 [1.369, 1.706] & 0\% \\
\Square{}, five checkpoints, 25\,s & 0.18 & 1.565 [1.457, 1.691] & 0\% & 1.600 [1.515, 1.693] & 0\% \\
\Square{}, five checkpoints, 25\,s & 0.32 & 1.713 [1.621, 1.804] & 5\% & 1.640 [1.566, 1.712] & 5\% \\
\Square{}, final checkpoint, 100\,s & 0.07 & 1.098 [0.971, 1.294] & 0\% & 1.040 [0.894, 1.226] & 0\% \\
\Square{}, final checkpoint, 100\,s & 0.28 & 1.129 [1.049, 1.213] & 0\% & 1.138 [1.070, 1.209] & 0\% \\
\Square{}, final checkpoint, 100\,s & 0.50 & 1.152 [1.054, 1.264] & 4\% & 1.195 [1.096, 1.269] & 13\% \\
\Can{}, five checkpoints, 25\,s & 0.15 & 1.624 [1.423, 1.839] & 4\% & 1.638 [1.453, 1.828] & 4\% \\
\Can{}, five checkpoints, 25\,s & 0.41 & 1.653 [1.413, 1.986] & 0\% & 1.908 [1.782, 2.061] & 0\% \\
\Can{}, five checkpoints, 25\,s & 0.67 & 2.176 [1.990, 2.414] & 0\% & 2.151 [2.020, 2.311] & 0\% \\
\Can{}, final checkpoint, 100\,s & 0.15 & 1.341 [1.161, 1.546] & 6\% & 1.347 [1.146, 1.553] & 6\% \\
\Can{}, final checkpoint, 100\,s & 0.45 & 1.383 [1.236, 1.639] & 0\% & 1.573 [1.462, 1.705] & 0\% \\
\Can{}, final checkpoint, 100\,s & 0.74 & 1.769 [1.623, 1.946] & 0\% & 1.793 [1.627, 1.944] & 0\% \\
\bottomrule
\end{tabular}
\caption{\textbf{All-\Worse{} multiples from inverting each set's own success curve.} Three target success rates $u$ near the bottom, middle, and top of the range that all sets reach. Brackets: percentile intervals from 20000 paired whole-seed draws that have solutions within the tested sizes. No-solution percentages include draws without a common range.}
\label{tab:equal-success-targets}
\end{table}

\begin{table}[!ht]
\centering\fontsize{8}{10}\selectfont
\setlength{\tabcolsep}{3pt}
\begin{tabular}{@{}cccc@{}}
\toprule
$u$ & $k(1,u)$ & $k(0.5,u)$ & Excess, \% \\
\midrule
\multicolumn{4}{@{}l}{\emph{Five-checkpoint evaluation, 25\,s}} \\
0.10 & 1.509\,[1.317, 1.692] & 1.425\,[1.288, 1.556] & 18\,[10, 26] \\
0.20 & 1.598\,[1.488, 1.723] & 1.387\,[1.277, 1.527] & 13\,[5, 23] \\
0.30 & 1.704\,[1.618, 1.802] & 1.492\,[1.406, 1.593] & 18\,[12, 26] \\
\midrule
\multicolumn{4}{@{}l}{\emph{Final-checkpoint evaluation, 25\,s}} \\
0.10 & 1.628\,[1.376, 1.869] & 1.423\,[1.230, 1.591] & 15\,[2, 28] \\
0.20 & 1.534\,[1.420, 1.674] & 1.368\,[1.250, 1.540] & 13\,[5, 25] \\
0.30 & 1.692\,[1.551, 1.808] & 1.464\,[1.346, 1.591] & 16\,[8, 27] \\
\midrule
\multicolumn{4}{@{}l}{\emph{Final-checkpoint evaluation, 100\,s}} \\
0.10 & 1.201\,[1.043, 1.340] & 1.196\,[1.051, 1.354] & 10\,[\ensuremath{-}0.4, 22] \\
0.20 & 1.116\,[0.990, 1.269] & 1.186\,[1.077, 1.347] & 12\,[1, 25] \\
0.30 & 1.142\,[1.062, 1.229] & 1.254\,[1.153, 1.394] & 18\,[8, 30] \\
\midrule
\multicolumn{4}{@{}l}{\emph{Final-checkpoint evaluation, 200\,s}} \\
0.10 & 1.126\,[0.986, 1.299] & 1.182\,[1.053, 1.336] & 12\,[2, 24] \\
0.20 & 1.062\,[0.929, 1.220] & 1.175\,[1.056, 1.352] & 14\,[3, 27] \\
0.30 & 1.084\,[1.003, 1.167] & 1.210\,[1.129, 1.342] & 16\,[8, 29] \\
\bottomrule
\end{tabular}
\space
\caption{\textbf{Excess of the half-\Better{}, half-\Worse{} set on \Square{} from monotone-curve inversion.} Targets lie within the common attainable range of the three sets. Brackets: percentile intervals from resampled seeds; seed draws are shared across sets, sizes, and limits. A multiple uses the draws in which its two curves have solutions, and an excess uses the draws in which all three do. Every draw has all three solutions except 0.005\% of the draws at $u=0.30$ in the five-checkpoint evaluation.}
\label{tab:equal-success-mixing}
\end{table}

\subsection{Multiple comparisons and initial-state resampling}
\label{app:multiplicity}
\Cref{tab:sensitivity} uses the final-checkpoint fits of \cref{tab:replay} for three \Square{} quantities.
They are the change in $\ln k(1)$ from 25\,s to 100\,s, the log excess $E(0.5)$ at 25\,s, and the change in $E(0.5)$ between the same limits.
The alternative fits keep only $p\in\{0, 0.5, 1\}$, or the first 12 seeds, which have all five sets.
The Bonferroni correction covers these three quantities.
The change in $\ln k(1)$ and the log excess at 25\,s exclude zero under every construction shown.
For the change in $E(0.5)$, the main seed interval lies within $\pm0.04$.
\Cref{tab:extended-uncertainty} resamples seeds and shared initial states jointly for the extension from 100\,s to 200\,s.

\begin{table}[!ht]
  \centering\small
  \setlength{\tabcolsep}{4pt}
\begin{tabular}{@{}lccc@{}}
\toprule
Fit / seeds & $\Delta\ln k(1)$, 25$\to$100\,s & $E(0.5)$, 25\,s & $\Delta E(0.5)$, 25$\to$100\,s \\
\midrule
All sets / 24& \ensuremath{-}0.33 [\ensuremath{-}0.40, \ensuremath{-}0.27] & 0.13 [0.04, 0.21] & 0.00 [\ensuremath{-}0.03, 0.03] \\
$p\in\{0, 0.5, 1\}$ / 24& \ensuremath{-}0.33 [\ensuremath{-}0.39, \ensuremath{-}0.27] & 0.12 [0.04, 0.21] & 0.00 [\ensuremath{-}0.03, 0.03] \\
All sets / first 12& \ensuremath{-}0.30 [\ensuremath{-}0.41, \ensuremath{-}0.19] & 0.10 [0.01, 0.19] & 0.00 [\ensuremath{-}0.05, 0.05] \\
$p\in\{0, 0.5, 1\}$ / first 12& \ensuremath{-}0.29 [\ensuremath{-}0.38, \ensuremath{-}0.20] & 0.10 [0.01, 0.18] & 0.00 [\ensuremath{-}0.05, 0.05] \\
\midrule
\multicolumn{4}{@{}l}{\emph{Main fit: percentile intervals by resampled unit}} \\
Seeds & [\ensuremath{-}0.39, \ensuremath{-}0.28] & [0.06, 0.20] & [\ensuremath{-}0.03, 0.03] \\
Shared initial states & [\ensuremath{-}0.38, \ensuremath{-}0.29] & [0.07, 0.18] & [\ensuremath{-}0.02, 0.03] \\
Seeds + shared initial states & [\ensuremath{-}0.42, \ensuremath{-}0.26] & [0.03, 0.23] & [\ensuremath{-}0.05, 0.05] \\
\bottomrule
\end{tabular}
\space
  \caption{\textbf{Uncertainty for three \Square{} quantities.} Log scale; changes pair seeds from 25\,s to 100\,s. Top: estimates and seed intervals. Bottom: percentile intervals of the main fit by resampled unit, 20{,}000 draws, with seed resampling stratified by set coverage. An initial-state draw resamples the 100 shared initial states once and applies the same draw to every seed, set, size, and limit; each policy keeps its final checkpoint. All intervals use Bonferroni coverage $1-0.05/3$ across the three quantities.}
  \label{tab:sensitivity}
\end{table}

\begin{table}[!ht]
  \centering\footnotesize
  \setlength{\tabcolsep}{3pt}
\begin{tabular}{@{}llccc@{}}
\toprule
Quantity & Task & 100\,s & 200\,s & Paired change \\
\midrule
$k(1)$ & \Square{} & [0.97, 1.27] & [0.95, 1.23] & [\ensuremath{-}0.06, \ensuremath{-}0.004] \\
$k(1)$ & \Can{} & [1.42, 1.86] & [1.39, 1.81] & [\ensuremath{-}0.04, \ensuremath{-}0.01] \\
$E(0.5)$ & \Square{} & [0.05, 0.21] & [0.06, 0.21] & [\ensuremath{-}0.02, 0.02] \\
\bottomrule
\end{tabular}
\space
  \caption{\textbf{Intervals at 100\,s and 200\,s with seeds and shared initial states resampled jointly.} Point estimates are in \cref{tab:replay}; brackets are percentile intervals. Changes are 200\,s minus 100\,s, in $\ln k(1)$ for the first two rows and in $E(0.5)$ for the last.}
  \label{tab:extended-uncertainty}
\end{table}
\section{Time-Limit Results}
\label{app:time-limit-details}

\subsection{Results across time limits}
\label{app:limits}
\Cref{tab:replay} reports the fits at the four time limits of the main text, and \cref{fig:horizon} shows the full curves.
The \Can{} fits include its half-\Better{}, half-\Worse{} set, which shares the slope with the other two sets.
The fractions of $\ln k(1)$ removed divide the paired mean reduction by the mean $\ln k(1)$ at 25\,s within the same seeds.
Their Fieller limits can exceed one because the denominator is itself uncertain.
Across the limits of \cref{fig:horizon}, every 5\,s from 20\,s to 200\,s, the excess of the half-\Better{}, half-\Worse{} set ranges from 13\% at 20\,s to 14\% at 45\,s.
The all-\Worse{} multiple changes little after about 150\,s.
From 150\,s to 200\,s, $k(1)$ stays at 1.08 on \Square{} and moves from 1.59 to 1.58 on \Can{}; $E(0.5)$ stays at 0.13 on \Square{}.
At $N=24$, the half-\Better{}, half-\Worse{} and all-\Worse{} sets reach nearly equal totals within 100\,s, 246 against 245 successes.
The half-\Better{}, half-\Worse{} set thus falls below the all-\Worse{} set at $N=100$ (\cref{sec:mixing}) but not at $N=24$.

\begin{table}[!ht]
\centering\footnotesize
\setlength{\tabcolsep}{2pt}
\begin{tabular}{@{}lccccc@{}}
\toprule
 & \multicolumn{3}{c}{\Square{}} & \multicolumn{2}{c}{\Can{}, 6 sizes} \\
Limit (s) & $k(0.5)$ & $k(1)$ & $E(0.5)$ & $k(1)$ & $E(0.5)$ \\
\midrule
25& 1.37 [1.26, 1.49]& 1.55 [1.39, 1.73]& 0.13 [0.06, 0.19]& 1.88 [1.72, 2.07]& 0.09 [0.01, 0.16]\\
40& 1.28 [1.18, 1.39]& 1.30 [1.18, 1.43]& 0.13 [0.07, 0.19]& 1.72 [1.58, 1.90]& 0.09 [0.02, 0.16]\\
100& 1.19 [1.10, 1.29]& 1.11 [1.02, 1.22]& 0.13 [0.07, 0.19]& 1.61 [1.48, 1.77]& 0.09 [0.01, 0.16]\\
200& 1.18 [1.10, 1.26] & 1.08 [0.99, 1.18] & 0.13 [0.07, 0.19] & 1.58 [1.45, 1.73] & 0.09 [0.02, 0.17] \\
\bottomrule
\end{tabular}
\par\smallskip
\begin{tabular}{@{}lccc@{}}
\toprule
\Square{} limit (s) & Mean slope difference & $\ln k(1)-\ln k(0.5)$ & $E(0.75)$ \\
\midrule
25& \ensuremath{-}0.20 [\ensuremath{-}0.36, \ensuremath{-}0.03] & 0.12 [0.04, 0.21] & 0.10 [0.03, 0.17]\\
40& \ensuremath{-}0.21 [\ensuremath{-}0.38, \ensuremath{-}0.04] & 0.02 [\ensuremath{-}0.06, 0.09] & 0.07 [\ensuremath{-}0.03, 0.17]\\
100& \ensuremath{-}0.18 [\ensuremath{-}0.36, \ensuremath{-}0.004] & \ensuremath{-}0.07 [\ensuremath{-}0.13, 0.002] & 0.06 [\ensuremath{-}0.04, 0.16]\\
\bottomrule
\end{tabular}
\par\smallskip
\begin{tabular}{@{}lcccc@{}}
\toprule
Paired change & $\Delta\ln k(0.5)$ & $\Delta\ln k(1)$ & $\Delta E(0.5)$ & $\Delta$ slope \\
\midrule
\Square{}, 100$-$25\,s & \ensuremath{-}0.14 [\ensuremath{-}0.16, \ensuremath{-}0.12]& \ensuremath{-}0.33 [\ensuremath{-}0.38, \ensuremath{-}0.28]& 0.00 [\ensuremath{-}0.03, 0.03]& 0.01 [\ensuremath{-}0.05, 0.07] \\
\Square{}, 200$-$100\,s & \ensuremath{-}0.01 [\ensuremath{-}0.03, 0.01] & \ensuremath{-}0.03 [\ensuremath{-}0.05, \ensuremath{-}0.01] & 0.00 [\ensuremath{-}0.01, 0.02] & -- \\
\Can{}, 100$-$25\,s & -- & \ensuremath{-}0.15 [\ensuremath{-}0.18, \ensuremath{-}0.13]& -- & -- \\
\Can{}, 200$-$100\,s & -- & \ensuremath{-}0.02 [\ensuremath{-}0.03, \ensuremath{-}0.02] & -- & -- \\
\bottomrule
\end{tabular}
\par\smallskip
\begin{tabular}{@{}lc@{}}
\toprule
Fit with $p\in\{0, 0.5, 1\}$, 25$\to$100\,s & Fraction of $\ln k(1)$ removed \\
\midrule
\Square{}, all sizes & 75\% [62\%, 94\%]\\
\Square{}, $N=24,100$, first 12 seeds & 80\% [62\%, 121\%]\\
\Can{}, all sizes & 24\% [19\%, 30\%]\\
\bottomrule
\end{tabular}
\space
\caption{\textbf{Results across time limits.} All sizes and sets, 100 episodes per policy; $E$: log excess. Brackets: seed intervals for the \Square{} multiples, log excesses, and $\ln k(1)-\ln k(0.5)$; stratified percentile intervals for the mean slope difference; whole-seed percentile intervals on \Can{}. Changes pair seeds, and removed fractions use paired Fieller intervals.}
\label{tab:replay}
\end{table}

\subsection{Five-checkpoint evaluation}
\label{app:five-checkpoint}
The five-checkpoint evaluation combines the last five checkpoints of every policy (\cref{tab:protocols}) and gives multiples and log excesses of similar size at 25\,s (\cref{tab:checkpoint-summary}).
Across its time limits from 15\,s to 40\,s, every 2.5\,s, the log excess $E$ of the half-\Better{}, half-\Worse{} set lies between 0.12 and 0.15 and that of the 75\%-\Worse{} set between 0.10 and 0.13, with intervals that exclude zero.

\begin{table}[!ht]
\centering\small
\setlength{\tabcolsep}{5pt}
\begin{tabular}{@{}llccc@{}}
\toprule
Task & \Worse{} share $p$ & Seeds & $k(p)$ & $E(p)$ \\
\midrule
\Square{} & 0.25& 12& 1.17 [1.07, 1.27]& 0.05 [\ensuremath{-}0.02, 0.13]\\
 & 0.5& 24& 1.41 [1.33, 1.50]& 0.14 [0.10, 0.19]\\
 & 0.75& 12& 1.58 [1.41, 1.77]& 0.12 [0.07, 0.17]\\
 & 1 & 24& 1.60 [1.47, 1.73]& -- \\
\Can{} & 1 & 12& 1.95 [1.76, 2.16]& -- \\
\bottomrule
\end{tabular}
\space
\caption{\textbf{Five-checkpoint multiples and log excesses at 25\,s.} All sets; seed intervals. The 25\%- and 75\%-\Worse{} fits use the first 12 seeds and the all-\Better{} baseline of those seeds.}
\label{tab:checkpoint-summary}
\end{table}

\subsection{Completion events and behavior}
\label{app:events}
Step traces record the gripper--object contact and the object position at every step of an evaluation episode.
On \Square{}, they cover $p\in\{0, 0.5, 1\}$ at $N=24,100$.
Contact breaks shorter than 5 steps are merged, and a sustained grasp is a contact that then lasts at least 5 steps.
A counted grasp loss lasts at least 5 steps and starts with the object away from the goal and more than 1\,cm above its reset height.
We exclude terminal releases within 5 steps before success when contact never resumes.
A one-grasp success has one sustained grasp and no counted loss; an interrupted success has another grasp or a counted loss.
Time ratios pool the one-grasp successes and count the first grasp step.

Late completions are of both kinds (\cref{tab:square-events}).
\Cref{tab:timing-details} gives their percentage of the successes within 100\,s and the \Worse{}/\Better{} time ratios.
Over $N=24$ and 100, the two-size $\ln k(1)$ falls from 0.53 [0.40, 0.67] at 25\,s to 0.16 [0.05, 0.26] at 100\,s (24 seeds).
To split this fall, we add the late successes of one kind to the successes within 25\,s and then add those of the other kind.
After each step, we refit the shared slope of the sets with $p\in\{0, 0.5, 1\}$.
Each kind's contribution is its mean over the two orders of addition.
One-grasp successes account for 51\% [45\%, 57\%] of the fall, and interrupted successes for the rest.
The split depends on how contact events are defined: across the four rules of \cref{tab:event-rules}, one-grasp successes account for 8\% to 53\% of the fall.
\Cref{tab:completion-mixture} compares the half-\Better{}, half-\Worse{} set with the mean of the all-\Better{} and all-\Worse{} sets.
The comparison uses the unconditional cumulative success rate $S_p(t)$, the fraction of all episodes that succeed by time $t$.
Failures within 100\,s thus stay in the denominator of $S_{0.5}(t)-[S_0(t)+S_1(t)]/2$.
A simultaneous band over time at each size uses the 95th percentile of the centered supremum deviations across paired bootstrap draws.

\begin{table}[!ht]
\centering\small
\setlength{\tabcolsep}{3pt}

\begin{tabular}{@{}lccccc@{}}
\toprule
$N$ & $p$ & \shortstack{Late,\\interrupted/total} & \shortstack{Late,\\interrupted \%} & \shortstack{Early,\\interrupted/total} & \shortstack{Early,\\interrupted \%} \\
\midrule
24 & 0 & 28/33 & 85 [73, 95] & 57/294 & 19 [16, 23] \\
24 & 0.5 & 43/61 & 70 [59, 82] & 36/185 & 19 [15, 24] \\
24 & 1 & 63/103 & 61 [54, 69] & 27/142 & 19 [13, 26] \\
100 & 0 & 38/55 & 69 [57, 81] & 219/1308 & 17 [14, 19] \\
100 & 0.5 & 133/211 & 63 [55, 71] & 157/915 & 17 [15, 20] \\
100 & 1 & 210/412 & 51 [45, 57] & 129/841 & 15 [13, 18] \\
\bottomrule
\end{tabular}
\caption{\textbf{Interrupted successes of \Square{} policies.} Early and late: successes before and after 25\,s, within 100\,s. Counts sum over seeds. Brackets: paired whole-seed percentile intervals.}
\label{tab:square-events}
\end{table}

\begin{table}[!ht]
\centering\small
\setlength{\tabcolsep}{3pt}
\begin{tabular}{@{}p{0.60\linewidth}cc@{}}
    \toprule
    Quantity & \Better{} & \Worse{} \\
    \midrule
    \Square{} median success time (s), five-checkpoint evaluation & 9.3& 17.0\\
    \Square{} late completions at $N=100$, \% of successes within 100\,s & 4& 33\\
    \Can{} late completions, \% of successes within 100\,s & 2& 14\\
    \midrule
    \Square{} \Worse{}/\Better{} time ratios & \multicolumn{2}{c}{Estimate} \\
    Policies, grasp to success & \multicolumn{2}{c}{1.95 [1.84, 2.06]} \\
    Demonstrations, grasp to success & \multicolumn{2}{c}{1.74 [1.57, 1.91]} \\
    Policies, whole episode & \multicolumn{2}{c}{2.19 [2.04, 2.32]} \\
    \bottomrule
  \end{tabular}
\caption{\textbf{Completion timing on \Square{} and \Can{}.} Late: successes after 25\,s and within 100\,s. The \Square{} median uses the five-checkpoint evaluation and conditions on success within 25\,s at $N=100$; \Can{} sums over all sizes and seeds; time ratios use one-grasp successes at $N=100$. Brackets: paired whole-seed percentile intervals.}
\label{tab:timing-details}
\end{table}

\begin{table}[!ht]
\centering\small
\setlength{\tabcolsep}{4pt}
\begin{tabular}{@{}lc@{}}
\toprule
Event rule & One-grasp fraction, \% \\
\midrule
Raw contact losses; no release exclusion & 8 [4, 12] \\
Raw contacts; releases within 5 steps excluded & 40 [34, 47] \\
Contact breaks under 5 steps merged; releases within 5 steps excluded & 51 [45, 57] \\
Contact breaks under 5 steps merged; releases within 20 steps excluded & 53 [47, 59] \\
\bottomrule
\end{tabular}
\space
\caption{\textbf{One-grasp fraction of the fall in the two-size $\ln k(1)$ under four contact-event rules.} \Square{}, $N=24,100$, from 25\,s to 100\,s. The third rule is the one used in the text and in \cref{tab:square-events}; the interrupted fraction is the complement. Endpoint success counts and the total fall do not depend on the rule. Brackets: paired whole-seed percentile intervals.}
\label{tab:event-rules}
\end{table}

\begin{table}[!ht]
\centering\small
\setlength{\tabcolsep}{3pt}

\begin{tabular}{@{}lcc@{}}
\toprule
Half-\Better{}, half-\Worse{} minus all-\Worse{}, $N=100$ & 25\,s & 100\,s \\
\midrule
Success difference, percentage points & 3 [\ensuremath{-}1, 7] & \ensuremath{-}5 [\ensuremath{-}9, \ensuremath{-}1] \\
\bottomrule
\end{tabular}\par\smallskip
\begin{tabular}{@{}lc@{}}
\toprule
Late contribution to that difference & Percentage points \\
\midrule
Additional late completions & \ensuremath{-}8 [\ensuremath{-}11, \ensuremath{-}6] \\
Late, one grasp & \ensuremath{-}5 [\ensuremath{-}7, \ensuremath{-}4] \\
Late, interrupted & \ensuremath{-}3 [\ensuremath{-}5, \ensuremath{-}2] \\
\bottomrule
\end{tabular}\par\smallskip
\begin{tabular}{@{}lcccc@{}}
\toprule
$N$ & \shortstack{Difference within\\100\,s, pp} & \shortstack{Maximum absolute\\difference, pp} & \shortstack{Simultaneous\\half-width, pp} & \shortstack{Mean duration\\difference, s} \\
\midrule
24 & \ensuremath{-}2 [\ensuremath{-}3, 0.06] & 2 & 2 & 1.40 [0.04, 2.76] \\
100 & \ensuremath{-}8 [\ensuremath{-}11, \ensuremath{-}4] & 8 & 4 & 6.33 [3.60, 9.26] \\
\bottomrule
\end{tabular}\caption{\textbf{Half-\Better{}, half-\Worse{} set against the all-\Worse{} set and against the mean curve on \Square{}.} Top: success differences from the all-\Worse{} set over all episodes at $N=100$, and the late successes that make up the change in this difference from 25\,s to 100\,s. Bottom: the half-\Better{}, half-\Worse{} curve minus the mean of the $p=0$ and $p=1$ curves, with durations censored at 100\,s; the band is simultaneous over time. Brackets: paired whole-seed percentile intervals; \cref{fig:completion} uses seed intervals instead.}
\label{tab:completion-mixture}
\end{table}

On \Square{}, \Worse{} demonstrations regrasp more often than \Better{} ones (\cref{tab:demo-events}), and on both tasks all-\Worse{} policies pause and regrasp more often than all-\Better{} policies (\cref{tab:policy-features}).
Both tables use the same contact rules.
Motion features use the end-effector displacement per step multiplied by the control frequency.
A pause requires at least 10 consecutive steps below 0.02\,m/s, and pause percentages average trajectory-level fractions.
The observation window ends at first success, or at 100\,s for an unsuccessful policy episode.
A regrasp means at least two sustained grasps, including episodes that end in failure; a loss then regrasp requires a counted grasp loss followed by another sustained grasp within the window.

\begin{table}[!ht]
\centering\footnotesize
\setlength{\tabcolsep}{2pt}
\begin{tabular}{@{}lcccccc@{}}
\toprule
Task/teleoperator & \shortstack{Pause\\\%} & \shortstack{Interrupted/\\total} & \shortstack{Interrupted\\\%} & \shortstack{Regrasp\\count (\%)} & \shortstack{Grasp\\loss} & \shortstack{Loss then\\regrasp, \%} \\
\midrule
\Square{}/B1 & $5$ & 6/50 & 12 [4, 22] & 1 ($2$) & 5 & $0$ \\
\Square{}/B2 & $2$ & 3/50 & 6 [0, 14] & 1 ($2$) & 3 & $2$ \\
\Square{}/W1 & $5$ & 24/50 & 48 [34, 62] & 10 ($20$) & 20 & $8$ \\
\Square{}/W2 & $11$ & 14/50 & 28 [16, 40] & 9 ($18$) & 10 & $6$ \\
\Can{}/B1 & $0$ & -- & -- & -- & -- & -- \\
\Can{}/B2 & $0$ & -- & -- & -- & -- & -- \\
\Can{}/W1 & $1$ & -- & -- & -- & -- & -- \\
\Can{}/W2 & $9$ & -- & -- & -- & -- & -- \\
\bottomrule
\end{tabular}
\space
\caption{\textbf{Demonstration events and pauses.} 50 demonstrations per teleoperator, through first success. A regrasp can overlap with a counted grasp loss. Interrupted-rate brackets resample demonstrations within each teleoperator; other features are point estimates. The \Can{} data lack contact fields (dashes).}
\label{tab:demo-events}
\end{table}

\begin{table}[!ht]
\centering\footnotesize
\setlength{\tabcolsep}{3pt}
\begin{tabular}{@{}lccccc@{}}
\toprule
Task & $N$ & $p$ & Pause, \% & Regrasp, \% & Loss then regrasp, \% \\
\midrule
\Square{} & 24 & 0 & $7\,[6, 8]$ & $12\,[11, 14]$ & $4\,[3, 5]$ \\
\Square{} & 24 & 0.5 & $11\,[9, 13]$ & $19\,[17, 21]$ & $4\,[3, 4]$ \\
\Square{} & 24 & 1 & $16\,[14, 17]$ & $25\,[23, 27]$ & $6\,[4, 7]$ \\
\Square{} & 100 & 0 & $8\,[7, 9]$ & $9\,[8, 10]$ & $4\,[4, 5]$ \\
\Square{} & 100 & 0.5 & $16\,[15, 17]$ & $22\,[21, 24]$ & $7\,[6, 8]$ \\
\Square{} & 100 & 1 & $22\,[21, 24]$ & $27\,[25, 29]$ & $7\,[6, 8]$ \\
\Can{} & 26 & 0 & $9\,[7, 12]$ & $2\,[1, 3]$ & $1\,[0, 2]$ \\
\Can{} & 26 & 1 & $18\,[13, 23]$ & $8\,[6, 11]$ & $4\,[3, 6]$ \\
\Can{} & 100 & 0 & $1\,[1, 1]$ & $1\,[0, 1]$ & $0\,[0, 1]$ \\
\Can{} & 100 & 1 & $8\,[7, 9]$ & $7\,[6, 8]$ & $4\,[3, 5]$ \\
\bottomrule
\end{tabular}
\space
\caption{\textbf{Policy pauses and regrasping.} Within 100\,s, all episodes including failures; \Square{}: 24 seeds, \Can{}: 12, 100 initial states per condition. Brackets: percentile intervals from 20{,}000 whole-seed draws, paired across sizes and sets within each task.}
\label{tab:policy-features}
\end{table}

\subsection{\texorpdfstring{\Can{}}{Can}}
\label{app:can-completion-times}
\Can{} traces use the same event rules at $N=26,100$ and 12 seeds, with entry into the target bin in place of nut placement (\cref{tab:can-time-limits}).
All-\Worse{} \Can{} policies take longer from grasp to success than all-\Better{} ones, and late one-grasp and interrupted successes both contribute to the fall of the two-size $\ln k(1)$ from 25\,s to 100\,s.
The \Can{} all-\Worse{} line is shallower than the all-\Better{} line, so its multiple also grows with $N$.

\begin{table}[!ht]
\centering\small
\setlength{\tabcolsep}{3pt}
\begin{tabular}{@{}lccc@{}}
\toprule
\Can{} $p$ & Success at $N=100$ & $\beta_p$ & $\beta_p-\beta_0$ \\
\midrule
0 & 0.93& 2.02& -- \\
1 & 0.69& 1.64& \ensuremath{-}0.38 [\ensuremath{-}0.59, \ensuremath{-}0.16]\\
\bottomrule
\end{tabular}\par\smallskip
\begin{tabular}{@{}lccccc@{}}
    \toprule
    $N$ & $p$ & Late, interrupted/total & Late, \% & Early, interrupted/total & Early, \% \\
    \midrule
    26& 1 & 43/83& 52 [44, 60]& 63/275& 23 [17, 31]\\
    100& 1 & 40/83& 48 [38, 58]& 154/847& 18 [16, 20]\\
    \bottomrule
  \end{tabular}\par\smallskip
\begin{tabular}{@{}p{0.53\linewidth}l@{}}
\toprule
\Can{} summary & Estimate\\
\midrule
Grasp-to-success time ratio, $N=26$ / $100$ & 2.24 [1.99, 2.60] / 1.88 [1.77, 2.05]\\
Two-size $\ln k(1)$, 25\,s / 100\,s & 0.74 [0.60, 0.88] / 0.57 [0.44, 0.71]\\
Paired reduction; one-grasp contribution & 0.17 [0.12, 0.21]; 0.08 [0.05, 0.11]\\
Interrupted contribution; one-grasp fraction & 0.09 [0.06, 0.12]; 48\% [36\%, 58\%]\\
\bottomrule
\end{tabular}
\caption{\textbf{\Can{} slopes and completion events.} Top: five-checkpoint evaluation at 25\,s, 12 seeds. Middle: traces of the all-\Worse{} set, early and late as in \cref{tab:square-events}. Bottom: comparisons of the all-\Worse{} with the all-\Better{} set. Brackets: whole-seed percentile intervals for interrupted fractions, the one-grasp fraction, and time ratios; seed intervals for the slope difference, two-size changes, and contributions.}
\label{tab:can-time-limits}
\end{table}
\section{The Excess: Training Samples, Sampler, and Teleoperators}
\label{app:excess}

\subsection{Training-sample accounting}
\label{app:sample-share}
The default sampler draws each demonstration in proportion to its length.
Crediting the groups by sequence starts then gives two predictions.
The first values sequence starts and converts the value back to demonstrations.
A \Worse{} sequence start is worth $1/[r\,k(1)]$ \Better{} sequence starts, where $r$ is the \Worse{}/\Better{} length ratio.
A \Worse{} demonstration holds $r$ times as many sequence starts as a \Better{} one on average, so it is worth $r/[r\,k(1)]=1/k(1)$ \Better{} demonstrations.
The ratio $r$ cancels, and the result is again the additive prediction by demonstration counts.
Counting each set's own sequence starts in place of the mean ratio $r$ changes the five-checkpoint log excess at 25\,s only from $E=0.14$ to 0.15.

The second changes the reference prediction.
The training-sample share prediction credits each group by its share $q$ of training sequences instead of its share $p$ of demonstrations, $1/\kadd(q)=(1-q)+q/k(1)$.
The expected \Worse{} shares of sequence starts are 0.39, 0.66, and 0.85 for $p=0.25,0.5,0.75$.
The log excess over this prediction averages $\ln k_s(p)-\ln\kadd(q_{sN};k_s(1))$ over sizes, where $q_{sN}$ is the \Worse{} share of sequence starts at size $N$.
Under the demonstration-uniform sampler of \cref{app:sampling}, $q=p$, so this prediction equals the additive prediction.
For the half-\Better{}, half-\Worse{} set in the five-checkpoint evaluation at 25\,s, the log excess over the training-sample share prediction is $E=0.07$ [0.02, 0.12], against $E=0.14$ over the additive prediction (\cref{tab:sample-shares}).
\Cref{tab:sample-shares} gives the ratio of these two log excesses in each evaluation, and the same ratio when every sequence is credited to \Worse{} demonstrations.
The five-checkpoint fits use all five sets; the final-checkpoint fits use $p\in\{0, 0.5, 1\}$.
Any prediction that reweights the two groups lies between 1 and $k(1)$, so reweighting alone cannot produce the ordering $k(0.5)>k(1)$ of the point estimates at 100\,s.

\begin{table}[!ht]
\centering\small
\setlength{\tabcolsep}{3pt}
\begin{tabular}{@{}lccc@{}}
\toprule
Evaluation & $p$ & \shortstack{Log excess over the training-\\sample share prediction, $E$} & \shortstack{Ratio to the log excess over\\the additive prediction}\\
\midrule
Five checkpoints, 25\,s & 0.25 & \ensuremath{-}0.01 [\ensuremath{-}0.08, 0.07]& -- \\
Five checkpoints, 25\,s & 0.5 & 0.07 [0.02, 0.12]& 48\% [22\%, 62\%]\\
Five checkpoints, 25\,s & 0.75 & 0.06 [0.01, 0.11]& 51\% [11\%, 72\%]\\
Final checkpoint, 25\,s & 0.5 & -- & 43\% [\ensuremath{-}16\%, 64\%]\\
Final checkpoint, 40\,s & 0.5 & -- & 67\% [36\%, 80\%]\\
Final checkpoint, 100\,s & 0.5 & -- & 86\% [70\%, 97\%]\\
Final checkpoint, every sequence \Worse{}, 100\,s & 0.5 & -- & 51\% [\ensuremath{-}4\%, 88\%]\\
\bottomrule
\end{tabular}
\caption{\textbf{Log excess on \Square{} over the training-sample share prediction.} Ratios divide the seed-mean log excess over this prediction by the seed-mean log excess over the additive prediction from the same fits; the last row credits every sequence to \Worse{} demonstrations. Brackets: seed intervals for $E$ and paired Fieller intervals for the ratios.}
\label{tab:sample-shares}
\end{table}

\subsection{Sampler experiment}
\label{app:sampling}
The sampler experiment trains the all-\Better{} and all-\Worse{} sets and the half-\Better{}, half-\Worse{} set on \Square{} at the eight sizes of the main experiment.
It uses 12 of the main experiment's seeds and the same nested training sets.
Each set and size is trained once with the default sampler and once with the demonstration-uniform sampler, 576 policies in all.
The demonstration-uniform sampler draws a demonstration with probability $1/N$, with replacement, and then a start position uniformly within it.
The default sampler is robomimic's, unchanged.
The training data, recipe, and update budget are those of the main experiment.
All policies of a seed are trained and evaluated on one CPU model.
The final checkpoint of each policy is evaluated from the 100 shared initial states of the main experiment; each episode stops at its first success or after 200\,s.

\begin{figure}[!ht]
\centering
\includegraphics[width=\linewidth]{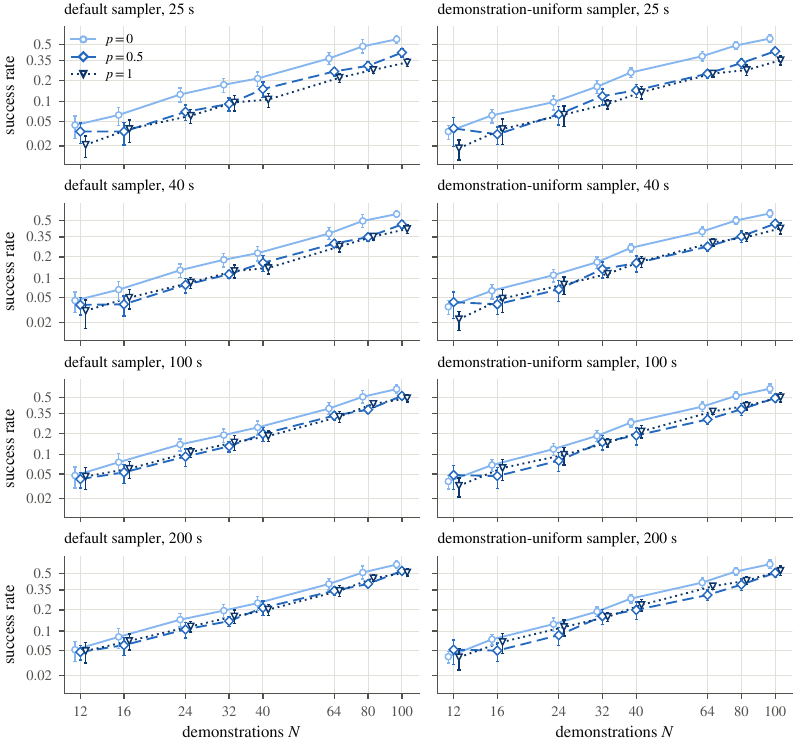}
\caption{\textbf{Success rates in the sampler experiment on \Square{}.} Rows: time limits; columns: samplers. Points: means over 12 seeds with seed intervals, offset along $N$ for legibility; lines connect the means. The vertical axis uses the log-odds scale.}
\label{fig:sampler-success}
\end{figure}

For each sampler and seed, the three sets share one slope: the mean of their three least-squares slopes.
Each sampler's slope is thus fitted on that sampler's own policies.
The multiples, the additive prediction, and the excess then follow \cref{sec:setup} (\cref{tab:sampler-fits}).
Under both samplers, the excess interval lies above zero at every limit, and the all-\Worse{} multiple falls from 25\,s to 100\,s.
\Cref{fig:sampler-success} shows the success rates of all sets and sizes.
Relative to the all-\Better{} set, the slope differences of the two sets that contain \Worse{} demonstrations have negative point estimates under both samplers at every limit, and all their intervals include zero (\cref{tab:sampler-slopes}).

\begin{table}[!ht]
\centering\footnotesize
\setlength{\tabcolsep}{3pt}
\begin{tabular}{@{}llccccc@{}}
\toprule
Sampler & Limit (s) & $k(1)$ & $k(0.5)$ & $\kadd(0.5)$ & $E(0.5)$ & Excess (\%) \\
\midrule
Default & 25& 1.64\,[1.46, 1.85] & 1.43\,[1.25, 1.63] & 1.24\,[1.19, 1.30] & 0.14\,[0.04, 0.25] & 15\,[4, 28] \\
Default & 40& 1.37\,[1.23, 1.53] & 1.32\,[1.16, 1.51] & 1.16\,[1.10, 1.21] & 0.14\,[0.04, 0.23] & 15\,[4, 26] \\
Default & 100& 1.18\,[1.06, 1.32] & 1.23\,[1.08, 1.41] & 1.08\,[1.03, 1.14] & 0.13\,[0.04, 0.23] & 14\,[4, 25] \\
Default & 200& 1.15\,[1.03, 1.28] & 1.20\,[1.05, 1.36] & 1.07\,[1.01, 1.12] & 0.12\,[0.03, 0.20] & 12\,[3, 23] \\
\midrule
Uniform & 25& 1.56\,[1.41, 1.72] & 1.39\,[1.28, 1.50] & 1.22\,[1.17, 1.26] & 0.13\,[0.06, 0.21] & 14\,[6, 23] \\
Uniform & 40& 1.34\,[1.23, 1.47] & 1.32\,[1.22, 1.43] & 1.15\,[1.10, 1.19] & 0.14\,[0.07, 0.22] & 15\,[7, 24] \\
Uniform & 100& 1.19\,[1.09, 1.28] & 1.27\,[1.18, 1.36] & 1.09\,[1.05, 1.12] & 0.16\,[0.08, 0.23] & 17\,[9, 25] \\
Uniform & 200& 1.14\,[1.05, 1.23] & 1.25\,[1.16, 1.36] & 1.07\,[1.03, 1.10] & 0.17\,[0.09, 0.24] & 18\,[9, 27] \\
\midrule
Uniform $-$ default & 25& -- & -- & -- & \ensuremath{-}0.01\,[\ensuremath{-}0.12, 0.11] & \ensuremath{-}1\,[\ensuremath{-}12, 12] \\
Uniform $-$ default & 40& -- & -- & -- & 0.01\,[\ensuremath{-}0.11, 0.12] & 1\,[\ensuremath{-}10, 12] \\
Uniform $-$ default & 100& -- & -- & -- & 0.02\,[\ensuremath{-}0.08, 0.13] & 2\,[\ensuremath{-}8, 14] \\
Uniform $-$ default & 200& -- & -- & -- & 0.05\,[\ensuremath{-}0.05, 0.14] & 5\,[\ensuremath{-}5, 16] \\
\bottomrule
\end{tabular}
\space
\caption{\textbf{Multiples, additive prediction, and excess in the sampler experiment on \Square{}.} 12 seeds, 100 episodes per policy; $E$: log excess, the mean of the per-seed log ratios. Brackets: Student $t$ intervals over seeds; for $k(1)$ and $k(0.5)$ they are computed for $\ln k$ and exponentiated, and excess percentages apply $100[\exp(E)-1]$ to the estimate and to both endpoints. As in \cref{app:terms}, $\kadd(0.5)$ is computed from the summary $k(1)$, with its interval from the endpoints of the $k(1)$ interval. The uniform $-$ default rows pair seeds; their last column is $100[\exp(E_{\mathrm{uniform}}-E_{\mathrm{default}})-1]\%$, the change in the ratio of required to predicted demonstrations rather than a difference of the two excess percentages.}
\label{tab:sampler-fits}
\end{table}

\begin{table}[!ht]
\centering\footnotesize
\setlength{\tabcolsep}{3pt}
\begin{tabular}{@{}lcccc@{}}
\toprule
 & \multicolumn{2}{c}{Default sampler} & \multicolumn{2}{c}{Demonstration-uniform sampler} \\
Limit (s) & $\beta_{0.5}-\beta_0$ & $\beta_1-\beta_0$ & $\beta_{0.5}-\beta_0$ & $\beta_1-\beta_0$ \\
\midrule
25& \ensuremath{-}0.09\,[\ensuremath{-}0.26, 0.08] & \ensuremath{-}0.13\,[\ensuremath{-}0.37, 0.12] & \ensuremath{-}0.13\,[\ensuremath{-}0.36, 0.09] & \ensuremath{-}0.14\,[\ensuremath{-}0.35, 0.06] \\
40& \ensuremath{-}0.08\,[\ensuremath{-}0.26, 0.10] & \ensuremath{-}0.09\,[\ensuremath{-}0.36, 0.18] & \ensuremath{-}0.15\,[\ensuremath{-}0.34, 0.05] & \ensuremath{-}0.10\,[\ensuremath{-}0.26, 0.07] \\
100& \ensuremath{-}0.06\,[\ensuremath{-}0.27, 0.15] & \ensuremath{-}0.12\,[\ensuremath{-}0.35, 0.12] & \ensuremath{-}0.15\,[\ensuremath{-}0.34, 0.04] & \ensuremath{-}0.06\,[\ensuremath{-}0.24, 0.11] \\
200& \ensuremath{-}0.08\,[\ensuremath{-}0.27, 0.10] & \ensuremath{-}0.11\,[\ensuremath{-}0.32, 0.11] & \ensuremath{-}0.14\,[\ensuremath{-}0.33, 0.06] & \ensuremath{-}0.07\,[\ensuremath{-}0.24, 0.10] \\
\bottomrule
\end{tabular}
\space
\caption{\textbf{Slope differences in the sampler experiment.} Slope of the half-\Better{}, half-\Worse{} set ($\beta_{0.5}$) or the all-\Worse{} set ($\beta_1$) minus that of the all-\Better{} set ($\beta_0$), in log-odds per $\ln N$. Brackets: Student $t$ intervals over 12 seeds.}
\label{tab:sampler-slopes}
\end{table}

\FloatBarrier

\subsection{Single teleoperators and teleoperator pairs}
\label{app:teleoperators}
In the single-teleoperator and teleoperator-pair experiments, compared policies are trained on the same machine and GPU model, and the teleoperators change.
At a fixed size, pooling changes the number of teleoperators, which teleoperators appear, and the demonstrations per teleoperator.

\paragraph{Pooling within a group.}
At $N=40$, 12 seeds each train the four single teleoperators and the two pooled groups.
The single-teleoperator minus pooled log-odds contrast is $\Delta_s=(z_{1s}+z_{2s})/2-z_{\mathrm{pool},s}$ (\cref{tab:single-operator-controls}).
The comparison across sizes pairs 6 seeds trained on one machine.

\begin{table}[!ht]
\centering\small
\setlength{\tabcolsep}{3pt}
\begin{tabular}{@{}llcc@{}}
    \toprule
    Evaluation & Limit (s) & \Better{} & \Worse{} \\
    \midrule
    Five checkpoints & 25& 0.64 [0.39, 0.88]& 0.27 [0.07, 0.47]\\
    Five checkpoints & 40& 0.65 [0.41, 0.90]& 0.25 [0.08, 0.43]\\
    Final checkpoint & 25& 0.60 [0.20, 0.99]& 0.34 [\ensuremath{-}0.02, 0.70]\\
    Final checkpoint & 40& 0.62 [0.23, 1.02]& 0.23 [\ensuremath{-}0.03, 0.48]\\
    Final checkpoint & 100& 0.60 [0.19, 1.01]& 0.22 [\ensuremath{-}0.03, 0.46]\\
    \bottomrule
  \end{tabular}
\par\smallskip
\begin{tabular}{@{}lcc@{}}
    \toprule
    $N$ or trend & \Better{} & \Worse{} \\
    \midrule
    12& 0.30 [\ensuremath{-}0.11, 0.71]& 0.52 [\ensuremath{-}0.33, 1.36]\\
    24& 0.30 [\ensuremath{-}0.23, 0.83]& 0.24 [\ensuremath{-}0.05, 0.52]\\
    40& 0.53 [0.16, 0.89]& 0.15 [\ensuremath{-}0.03, 0.33]\\
    Slope in $\ln N$ & 0.18 [\ensuremath{-}0.36, 0.72]& \ensuremath{-}0.31 [\ensuremath{-}1.10, 0.48]\\
    \bottomrule
  \end{tabular}
\space
\caption{\textbf{Single-teleoperator minus pooled log-odds within each group.} Top: $N=40$, 12 paired seeds. Bottom: five-checkpoint evaluation at 25\,s across sizes, 6 paired seeds trained on one machine; the last row is the least-squares slope of $\Delta_s(N)$ against $\ln N$. Brackets: paired seed intervals.}
\label{tab:single-operator-controls}
\end{table}

\paragraph{Single-teleoperator baseline.}
For teleoperator $a$ trained alone, let $z_a$ be the mean success log-odds of 3 policies at $N=40$ (\cref{tab:operators}).
Let $\bar z_{\mathrm B}$ be the mean of $z_{\mathrm{B1}}$ and $z_{\mathrm{B2}}$, and let $\beta=1.58$ be the mean slope of the main experiment's fits with $p\in\{0, 0.5, 1\}$.
Define
\[
 k_a=\exp[(\bar z_{\mathrm B}-z_a)/\beta],\qquad
 V_p=\frac{1}{N}\sum_a\frac{n_a}{k_a},\qquad
 V_{\mathrm{BB}}=\frac{k_{\mathrm{B1}}^{-1}+k_{\mathrm{B2}}^{-1}}{2},
\]
where $n_a$ counts the demonstrations from $a$ in the set with \Worse{} share $p$.
The teleoperator multiple $k_a$ makes a demonstration from $a$ worth $1/k_a$ demonstrations of the average single \Better{} teleoperator.
The calibration size $N=40$ is the size at which the multiples $k_a$ are estimated.
At this size, the predicted log-odds are $\bar z_{\mathrm B}+\beta\ln V_p$.
Predicted minus observed log-odds, divided by $\beta$, is the single-teleoperator log excess.
At $N=64,100$, the single-teleoperator baseline reads the seed's pooled B1+B2 line at size $NV_p/V_{\mathrm{BB}}$.
It applies the multiples calibrated at $N=40$ to both sizes, and so charges each pair the log-odds difference between B1+B2 and its two single teleoperators.
The additive prediction instead reads the same line at size $N[(1-p)+p/k_s(1)]$; both log-odds differences are divided by the seed's slope.
Intervals add the independent seed-mean and calibration variances $\sigma_i^2$ by the delta method.
Their Welch--Satterthwaite degrees of freedom are $(\sum_i\sigma_i^2)^2/\sum_i(\sigma_i^4/\nu_i)$, where $\nu_i$ is the degrees of freedom of term $i$.

\begin{table}[!ht]
  \centering\small
  \setlength{\tabcolsep}{4pt}
  \begin{tabular}{@{}lcc@{}}
    \toprule
    Set & $E$ & Single-teleoperator log excess \\
    \midrule
    \multicolumn{3}{@{}l}{\emph{$N=40$, baseline: average single \Better{} teleoperator}} \\
    B1+B2 & -- & 0.34 [0.16, 0.52]\\
    W1+W2 & -- & 0.19 [0.03, 0.35]\\
    Four teleoperators, $p=0.25$ & -- & 0.42 [0.21, 0.64]\\
    Four teleoperators, $p=0.5$ & -- & 0.41 [0.29, 0.54]\\
    Four teleoperators, $p=0.75$ & -- & 0.37 [0.20, 0.55]\\
    \midrule
    \multicolumn{3}{@{}l}{\emph{$N=64,100$, $p=0.5$; baseline: pooled B1+B2}} \\
    B1+W1 & 0.28 [0.18, 0.39]& 0.09 [\ensuremath{-}0.08, 0.26]\\
    B2+W1 & 0.17 [0.10, 0.24]& 0.10 [\ensuremath{-}0.07, 0.27]\\
    B1+W2 & \ensuremath{-}0.01 [\ensuremath{-}0.11, 0.09]& 0.02 [\ensuremath{-}0.14, 0.19]\\
    B2+W2 & \ensuremath{-}0.23 [\ensuremath{-}0.32, \ensuremath{-}0.15]& \ensuremath{-}0.10 [\ensuremath{-}0.27, 0.06]\\
    Mean of pairs & 0.05 [\ensuremath{-}0.02, 0.12]& 0.03 [\ensuremath{-}0.06, 0.12]\\
    Four teleoperators & 0.18 [\ensuremath{-}0.01, 0.37]& 0.16 [\ensuremath{-}0.04, 0.37]\\
    Pairs $-$ four (paired) & \ensuremath{-}0.13 [\ensuremath{-}0.29, 0.03]& \ensuremath{-}0.13 [\ensuremath{-}0.30, 0.03]\\
    Four, all 24 seeds & -- & 0.14 [0.05, 0.24]\\
    \bottomrule
  \end{tabular}
  \caption{\textbf{Pooling relative to single teleoperators and to the two groups.} Five-checkpoint evaluation at 25\,s. $E$: log excess over the additive prediction. Intervals for the single-teleoperator log excess include calibration uncertainty. Top: seed counts of the main experiment. Bottom: 6 paired seeds unless marked.}
  \label{tab:pairs}
\end{table}

\paragraph{Teleoperator pairs on one machine.}
Across sizes, the four pairs are trained on one machine for 4 seeds at $N=24$ to 100.
This same-machine comparison measures the combined variation in $k_a$ and in the effect of pooling.
For seed $s$, the unnormalized single-teleoperator values are $\widetilde V_{as}=\exp[z_{as,40}/\bar\beta_s^{(3)}]$.
The single-teleoperator baseline then predicts $c_{s0}+\beta_{s0}\ln[N(\widetilde V_{\mathrm{Bi},s}+\widetilde V_{\mathrm{Wj},s})/(\widetilde V_{\mathrm{B1},s}+\widetilde V_{\mathrm{B2},s})]$.
Averaged over the four pairs, the single-teleoperator log excess has slope 0.05 [\ensuremath{-}0.22, 0.31] against $\ln N$.
For each pair and size, the interval of the single-teleoperator log excess includes zero, except for B2+W2 at $N=64$, where the estimate is \ensuremath{-}0.16 [\ensuremath{-}0.30, \ensuremath{-}0.03].
Apart from this case, the intervals include zero, so these comparisons detect no departure from the single-teleoperator baseline.

At $N=100$, \cref{tab:operator-pairs} compares the pairs of 12 seeds with the eight-size curves of the main experiment, with matched evaluation and time limit.
Refitting the same-machine all-\Better{} and all-\Worse{} policies at $N=40,100$ gives a second set of curves.
Against them, the mean pair log excess is 0.09 [0.03, 0.15] in the five-checkpoint evaluation at 25\,s and 0.12 [0.03, 0.20] in the final-checkpoint evaluation at 100\,s.
In the final-checkpoint evaluation, its change from 25\,s to 100\,s is 0.01 [\ensuremath{-}0.04, 0.05] against the main experiment's curves and 0.03 [\ensuremath{-}0.03, 0.10] against the refitted curves.
With either the main experiment's curves or the refitted curves, the mean pair log excess lies above zero in the five-checkpoint evaluation at 25\,s and in the final-checkpoint evaluation at 100\,s.
The two evaluations differ, so these levels are not a change across time limits; the paired change is the one reported above.

\begin{table}[!ht]
\centering\small
\setlength{\tabcolsep}{3pt}
\begin{tabular}{@{}lcc@{}}
\toprule
Pair & Five checkpoints, 25\,s & Final checkpoint, 100\,s\\
\midrule
B1+W1 & 0.26 [0.17, 0.36]& 0.29 [0.17, 0.41]\\
B1+W2 & 0.05 [\ensuremath{-}0.05, 0.15]& 0.10 [\ensuremath{-}0.02, 0.23]\\
B2+W1 & 0.26 [0.19, 0.33]& 0.33 [0.18, 0.48]\\
B2+W2 & \ensuremath{-}0.16 [\ensuremath{-}0.26, \ensuremath{-}0.06]& \ensuremath{-}0.15 [\ensuremath{-}0.34, 0.04]\\
Mean & 0.10 [0.02, 0.18]& 0.14 [0.01, 0.27]\\
\bottomrule
\end{tabular}
\caption{\textbf{Teleoperator pairs against the main experiment's curves.} Log excess over the additive prediction at $N=100$, 12 seeds; paired seed intervals. The final-checkpoint mean at 25\,s is 0.14 [0.02, 0.25].}
\label{tab:operator-pairs}
\end{table}

\subsection{Teleoperators and additions}
\label{app:additions}
\Cref{tab:operators} compares the six teleoperators at $N=40$.
It also reports the log-odds gain from each all-\Better{} set of size $n$ to the half-\Better{}, half-\Worse{} set of size $2n$ that contains it, that is, from adding $n$ \Worse{} demonstrations.

\begin{table}[!ht]
\centering\fontsize{8}{9.5}\selectfont
\setlength{\tabcolsep}{3pt}
\begin{tabular}{@{}lcccccc@{}}
    \toprule
    & B1 & B2 & O1 & O2 & W1 & W2 \\
    \midrule
    Success log-odds & \ensuremath{-}0.96& \ensuremath{-}0.70& \ensuremath{-}1.11& \ensuremath{-}1.56& \ensuremath{-}2.30& \ensuremath{-}1.37\\
    Median success time (s) & 7.6& 11.2& 16.8& 14.1& 18.5& 18.8\\
    Multiple $k_a$ & 1.09& 0.92& 1.20& 1.58& 2.53& 1.41\\
    \bottomrule
  \end{tabular}
\par\smallskip
\begin{tabular}{@{}lc@{}}
    \toprule
    Log-odds gain, $n\to2n$ & Estimate \\
    \midrule
    Final checkpoint, 100\,s, range over $n$ & 0.77 to 1.21\\
    Smallest lower interval bound over sizes and limits & 0.29\\
    \bottomrule
  \end{tabular}
\space
\caption{\textbf{\Square{} teleoperators and additions.} Top: $N=40$, 3 seeds, five-checkpoint evaluation; O1 and O2 are the \Okay{} teleoperators. Log-odds and multiples use success within 25\,s; median times are within-seed medians over the successes within 40\,s, averaged over seeds. Bottom: observed gains for the four additions ($12\to24$, $16\to32$, $32\to64$, and $40\to80$ demonstrations); the last row is the smallest lower bound of the paired seed intervals over these additions and over 25, 40, and 100\,s.}
\label{tab:operators}
\end{table}

\subsection{Training devices}
\label{app:devices}
To test whether the training machine matters, we retrain on a GPU machine the $p=0$ and $p=0.5$ \Square{} policies of 9 seeds first trained on a CPU machine, at $N\in\{16, 32, 64, 100\}$.
Seeds, training data, training budget, checkpoints, and evaluation initial states are unchanged.
The two runs use different versions of the training code, so the comparison measures machine and code version together.
The estimated change in success at 25\,s is \ensuremath{-}2 percentage points over all 9 seeds and \ensuremath{-}3 over the 6 seeds whose two runs are evaluated on the same machine (\cref{tab:device-effects}).
Across rows and limits, the change in the slope contrast $\beta_{0.5}-\beta_0$ lies between \ensuremath{-}0.21 and \ensuremath{-}0.11, and every interval includes zero.
\begin{table}[!ht]
\centering\fontsize{8}{9.5}\selectfont
\setlength{\tabcolsep}{3pt}
\begin{tabular}{@{}lccccc@{}}
\toprule
Policies compared & \shortstack{Seeds/\\policies} & \multicolumn{2}{c}{Success-rate change, pp} & \multicolumn{2}{c}{Slope-contrast change} \\
\cmidrule(lr){3-4}\cmidrule(l){5-6}
& & 25\,s & 40\,s & 25\,s & 40\,s \\
\midrule
All seeds, both sets & 9/72 & \ensuremath{-}2 [\ensuremath{-}3, 0.2] & \ensuremath{-}2 [\ensuremath{-}4, 0.4] & \ensuremath{-}0.16 [\ensuremath{-}0.46, 0.13] & \ensuremath{-}0.11 [\ensuremath{-}0.46, 0.23] \\
Same evaluation machine, both sets & 6/48 & \ensuremath{-}3 [\ensuremath{-}4, \ensuremath{-}1] & \ensuremath{-}3 [\ensuremath{-}4, \ensuremath{-}1] & \ensuremath{-}0.21 [\ensuremath{-}0.71, 0.29] & \ensuremath{-}0.12 [\ensuremath{-}0.72, 0.47] \\
Same evaluation machine, $p=0$ & 6/24 & \ensuremath{-}2 [\ensuremath{-}3, \ensuremath{-}1] & \ensuremath{-}2 [\ensuremath{-}3, \ensuremath{-}1] & -- & -- \\
Same evaluation machine, $p=0.5$ & 9/36 & \ensuremath{-}2 [\ensuremath{-}4, 1] & \ensuremath{-}2 [\ensuremath{-}5, 1] & -- & -- \\
\bottomrule
\end{tabular}
\space
\caption{\textbf{Retraining on a second machine.} Differences are retrained minus original. Same evaluation machine: seeds whose original and retrained policies of the row's sets are evaluated on the same machine. Policies: paired policies over seeds, sizes, and sets. Success-rate changes average the row's sets and sizes within each seed. The slope contrast is $(\beta_{0.5}-\beta_0)_{\mathrm{retrained}}-(\beta_{0.5}-\beta_0)_{\mathrm{original}}$, in log-odds per $\ln N$, and needs both sets (dashes). Brackets: paired seed intervals.}
\label{tab:device-effects}
\end{table}
\section{Other Training Conditions}
\label{app:training}

\subsection{Longer training and matched exposure}
\label{app:exposure}
\Cref{tab:training-budget} pairs seeds across two training budgets (\cref{tab:inventory}).
The 2000-epoch policies of the main experiment are evaluated at the 5 checkpoints of epochs 1600--2000 with 500 episodes per policy.
The 4000-epoch policies are evaluated at the 5 checkpoints of epochs 3600--4000 with 250 episodes per policy, 50 per checkpoint, from new initial states.
Under a fixed update budget, each transition of the longer \Worse{} demonstrations is sampled less often.
Exposure matching sets the number of \Worse{} updates to $U=\lfloor U_0M_{\mathrm W}/M_{\mathrm B}\rfloor$, where $U_0$ is the original update count and $M_{\mathrm W}$ and $M_{\mathrm B}$ count \Worse{} and \Better{} transitions.
The nominal exposure is $UbL/M_{\mathrm W}$, with batch size $b=100$, sequence length $L=10$, and $U_0=e\,100$ at epoch $e$.
Checkpoint labels and final-checkpoint initial states stay fixed.
Exposure matching also adds optimizer updates.

Doubling the training budget multiplies $k(1)$ by 1.02 [0.98, 1.06] and raises the log excess $E(0.5)$ at 25\,s by 0.04 [0.004, 0.07].
Exposure matching lowers $\ln k(1)$ by 0.07 [\ensuremath{-}0.05, 0.19] at 25\,s and by 0.05 [\ensuremath{-}0.10, 0.20] at 100\,s; both intervals include zero.

\begin{table}[!ht]
\centering\small
\setlength{\tabcolsep}{3pt}
\begin{tabular}{@{}lccccc@{}}
\toprule
Paired change & $k(0.5)$ ratio & $k(1)$ ratio & $E(0.5)$, 25\,s & $E(0.5)$, 20\,s & $E(0.5)$, 40\,s \\
\midrule
2000$\to$4000 epochs & 1.05 [1.01, 1.08]& 1.02 [0.98, 1.06]& 0.04 [0.004, 0.07]& 0.04 [0.01, 0.08]& 0.03 [\ensuremath{-}0.0004, 0.07]\\
\bottomrule
\end{tabular}
\par\smallskip
\begin{tabular}{@{}lcccc@{}}
\toprule
Evaluation & Default $k(1)$ & Matched $k(1)$ & Reduction, $\ln k$ & Fraction removed\\
\midrule
Five checkpoints, 25\,s & 1.67 [1.49, 1.87]& 1.55 [1.36, 1.77]& 0.07 [\ensuremath{-}0.05, 0.19]& 14\% [\ensuremath{-}11\%, 36\%]\\
Final checkpoint, 100\,s & 1.20 [1.08, 1.33]& 1.14 [0.93, 1.39]& 0.05 [\ensuremath{-}0.10, 0.20]& 27\% [\ensuremath{-}51\%, 167\%]\\
\bottomrule
\end{tabular}
\caption{\textbf{\Square{} training budget and transition exposure.} Top: five-checkpoint evaluation, fits over all sets and sizes; multiple ratios at 25\,s and changes in the log excess $E(0.5)$; paired whole-seed percentile intervals. Bottom: all-\Better{} and all-\Worse{} fits at $N=40,100$, 8 seeds trained on one machine and GPU model; reductions are default minus matched; seed intervals, and Fieller intervals for the fractions.}
\label{tab:training-budget}
\end{table}

\subsection{Diffusion policy on two \texorpdfstring{\Square{}}{Square} sizes}
\label{app:diffusion}
Both policy classes use 8 paired seeds, $p\in\{0, 0.5, 1\}$, and $N\in\{24,100\}$, separate from the main experiment.
Fits use these two sizes at epoch 2000 with the same 100 initial states.
The diffusion policy is a denoising diffusion probabilistic model (DDPM)~\citep{ho2020denoising} with exponential-moving-average evaluation weights.
It observes 2 steps, predicts 16 actions, and executes 8 environment steps before replanning.
The conditional one-dimensional U-Net uses channel widths 256/512/1024, a 256-dimensional diffusion embedding, kernel width 5, and 8 normalization groups.
AdamW uses learning rate \ensuremath{10^{-4}}, weight decay \ensuremath{10^{-6}}, and a half-cycle cosine schedule after 500 warm-up updates; the EMA power is 0.75.
Training and inference both use 100 denoising steps, with a squared-cosine noise schedule, noise prediction, and denoised-sample clipping.
Epochs, updates per epoch, batch size, disabled normalization and gradient clipping, and boundary padding match \cref{app:details}.

\Cref{tab:diffusion} gives the change in $\ln k(1)$ from 25\,s to 100\,s and the log excess of the half-\Better{}, half-\Worse{} set at both limits.
\Cref{tab:diffusion-success} gives the success rates of the all-\Better{} and all-\Worse{} sets.
The diffusion policy succeeds more often than BC-RNN-GMM in every condition of \cref{tab:diffusion-success}.
Its $\ln k(1)$ changes by \ensuremath{-}0.40 from 25\,s to 100\,s, against \ensuremath{-}0.31 for BC-RNN-GMM on the same seeds and sizes, so the all-\Worse{} multiple falls with the longer limit under both policy classes.

\begin{table}[!ht]
  \centering\small
  \setlength{\tabcolsep}{2pt}
\begin{tabular}{@{}lcc@{}}
\toprule
 & \multicolumn{2}{c}{8 seeds} \\
Metric & Diffusion & BC-RNN-GMM \\
\midrule
$\Delta\ln k(1)$ & \ensuremath{-}0.40\,[\ensuremath{-}0.49,\ensuremath{-}0.31] & \ensuremath{-}0.31\,[\ensuremath{-}0.37,\ensuremath{-}0.25] \\
\midrule
$E(0.5)$, 25\,s & 0.10\,[\ensuremath{-}0.09,0.30] & 0.12\,[\ensuremath{-}0.06,0.30] \\
$E(0.5)$, 100\,s & 0.07\,[\ensuremath{-}0.11,0.24] & 0.08\,[\ensuremath{-}0.06,0.23] \\
\bottomrule
\end{tabular}
\space
  \caption{\textbf{Change in the log all-\Worse{} multiple from 25\,s to 100\,s, and log excess of the half-\Better{}, half-\Worse{} set, for both policy classes.} $E$: log excess. Seed intervals, paired for the change.}
  \label{tab:diffusion}
\end{table}

\begin{table}[!ht]
\centering\footnotesize
\setlength{\tabcolsep}{4pt}
\begin{tabular}{@{}lllccc@{}}
\toprule
$N$ & $p$ & Limit (s) & Diffusion, \% & BC-RNN-GMM, \% & Diffusion $-$ BC-RNN-GMM, pp \\
\midrule
24 & 0 & 25 & $36.00\,[30.02, 41.98]$ & $10.50\,[6.79, 14.21]$ & $25.50\,[16.89, 34.11]$ \\
24 & 0 & 100 & $38.50\,[32.28, 44.72]$ & $11.75\,[7.76, 15.74]$ & $26.75\,[18.35, 35.15]$ \\
24 & 1 & 25 & $29.50\,[22.13, 36.87]$ & $6.12\,[4.06, 8.19]$ & $23.38\,[17.02, 29.73]$ \\
24 & 1 & 100 & $39.25\,[32.33, 46.17]$ & $8.88\,[6.71, 11.04]$ & $30.38\,[24.85, 35.90]$ \\
\midrule
100 & 0 & 25 & $75.12\,[71.74, 78.51]$ & $54.00\,[47.27, 60.73]$ & $21.12\,[14.20, 28.05]$ \\
100 & 0 & 100 & $76.38\,[73.15, 79.60]$ & $55.88\,[50.08, 61.67]$ & $20.50\,[14.88, 26.12]$ \\
100 & 1 & 25 & $62.12\,[59.70, 64.55]$ & $37.12\,[31.96, 42.29]$ & $25.00\,[18.43, 31.57]$ \\
100 & 1 & 100 & $75.25\,[72.36, 78.14]$ & $54.62\,[49.81, 59.44]$ & $20.62\,[15.01, 26.24]$ \\
\bottomrule
\end{tabular}
\space
\caption{\textbf{Success of diffusion and BC-RNN-GMM policies on \Square{}.} 8 paired seeds, 100 common initial states per policy. Success rates in percent, differences in percentage points (pp). Brackets: seed intervals, not clipped to $[0,100]$. The BC-RNN-GMM policies use the same seeds and sizes; training recipes and machines differ between the two policy classes.}
\label{tab:diffusion-success}
\end{table}
\section{Collection Costs and Throughput}
\label{app:costs}

\paragraph{Break-even pay ratios.}
For \Worse{} share $p$, the \Worse{} demonstrations per \Better{}-equivalent demonstration are $m=k(1)$ for replacing and $m=\mu(p)=p/[1/k(p)-(1-p)]$ for adding.
For $0<p\leq1$, adding gains \Better{}-equivalent demonstrations when $1/k(p)>1-p$.
Replacing pays for every demonstration in the new set, and adding treats the existing \Better{} demonstrations as a sunk cost.
Let $r$ be the \Worse{}/\Better{} ratio of mean recording durations and $t_B$ the mean \Better{} duration.
With recording time as the only cost, the break-even pay ratio is $1/(rm)$, so $\ln\theta=-\ln r-\ln m$.
Its 95\% intervals add, on this log scale, the variance across policy seeds and the analytic variance of each teleoperator's mean demonstration length, propagated by the delta method, with Welch--Satterthwaite degrees of freedom; the interval of $\mu$ maps that of $k(p)$.

At 25\,s, matching 40 \Better{} demonstrations takes about 62 \Worse{} demonstrations and 3 times the recording time.
Adding 40 \Worse{} demonstrations to an equally large \Better{} set gains 18.4 [13.5, 23.7] \Better{}-equivalent demonstrations, so each \Better{}-equivalent demonstration gained costs $\mu(0.5)=40/18.4$ \Worse{} demonstrations.
This ratio includes the excess as well as the gap between the groups.

\paragraph{Overheads.}
Extra collection times $o_Bt_B$ and $o_Wt_B$ per accepted demonstration include resets, discarded attempts, and amortized teleoperator training.
For hourly rates $c_B$ and $c_W$, write $\rho=c_W/c_B$, and let $f=(F_C-F_B)/(c_Bt_B)$ be the fixed cost of the candidate set minus that of the \Better{} set, which includes policy training and deployment but not the costs already charged as collection time.
At time limit $h$, define
\begin{equation}
 v_h(p)=\frac{1/k_h(p)-(1-p)}{p},\qquad
 \theta_h(p)=\frac{(1+o_B)v_h(p)-f/(Np)}{r+o_W}.
 \label{eq:asymmetric-cost}
\end{equation}
For a positive \Better{}-equivalent gain, recording \Worse{} demonstrations costs less when $\rho<\theta_h(p)$, with the fixed-cost difference held constant as $\rho$ varies.
Setting $o_B=o_W=o$ and $f=0$ gives $(1+o)/[m(r+o)]$.
\Cref{tab:joint-costs} reports ratios without overheads and one asymmetric example.

\subsection{Throughput}
\label{app:throughput}
\Cref{tab:throughput} divides total successes by execution time, counting each failure at the full limit, and reports successes per 50\,s of execution (1000 steps).
For $y_s$ successes among $R_s$ attempts and $T_s$ executed steps, let $\alpha$ multiply the step duration.
A per-attempt overhead $d$, success resets $e_+$, and failure recovery $e_-$ are measured in policy steps:
\begin{equation}
 T_{\mathrm{eff},s}=\alpha T_s+R_sd+y_se_+ +(R_s-y_s)e_-,\qquad
 Q_h(p)=\frac{1000\sum_s y_s}{\sum_s T_{\mathrm{eff},s}}.
 \label{eq:deployment-overhead}
\end{equation}
The parameters can differ between a candidate policy and the all-\Better{} baseline.
\Cref{tab:throughput} reports $Q_h(0)$, $Q_h(1)$, and their ratio.

\begin{table}[!ht]
\centering\small
\setlength{\tabcolsep}{3pt}
\begin{tabular}{@{}lccc@{}}
\toprule
Limit (s) & All-\Better{} & All-\Worse{} & \Worse{}/\Better{}\\
\midrule
25& 1.618 [1.514, 1.728] & 0.796 [0.735, 0.861] & 0.492 [0.441, 0.551] \\
100& 0.571 [0.526, 0.621] & 0.425 [0.389, 0.463] & 0.744 [0.652, 0.852] \\
\bottomrule
\end{tabular}
\space
\caption{\textbf{Successes per 50\,s of execution on \Square{}.} $Q_h(p)$ with $\alpha=1$ and $d=e_+=e_-=0$; $N=100$; successful and failed episodes both count. Brackets: percentile intervals from equal-sized whole-seed blocks resampled jointly across sets and sizes; both groups use the same seeds.}
\label{tab:throughput}
\end{table}

\begin{table}[!ht]
\centering\footnotesize
\setlength{\tabcolsep}{3pt}
\begin{tabular}{@{}llcc@{}}
\toprule
 & & \multicolumn{2}{c}{Break-even pay ratio} \\
\cmidrule(l){3-4}
Limit (s) & $(o_B,o_W)$ & Replace & Add \\
\midrule
25 & $(0,0)$ & 0.33 [0.29, 0.38] & 0.24 [0.18, 0.32] \\
40 & $(0,0)$ & 0.40 [0.35, 0.45] & 0.29 [0.23, 0.37] \\
100 & $(0,0)$ & 0.47 [0.41, 0.53] & 0.35 [0.28, 0.44] \\
100 & $(0,1)$ & 0.31 [0.28, 0.34] & 0.23 [0.19, 0.29] \\
\bottomrule
\end{tabular}
\space
\caption{\textbf{\Square{} break-even pay ratios with and without collection overheads.} All sets, 24 seeds; hypothetical $f=0$. The inputs $o_B$ and $o_W$ use units of the \Better{} recording duration. Brackets: delta method with Welch--Satterthwaite degrees of freedom.}
\label{tab:joint-costs}
\end{table}

\end{document}